\documentclass[3p,twocolumn]{elsarticle}
\usepackage{hyperref}
\journal{Journal}
\usepackage{color,soul}
\usepackage{multirow}
\usepackage{booktabs}
\usepackage{amssymb,amsmath,array,amsfonts}
\usepackage{amsthm}
\usepackage{subfigure}
\usepackage[table,xcdraw]{xcolor}

\definecolor{darkgreen}{rgb}{0,0.6,0.2}

\newcommand{\h}{\mathbf{h}}

\def\x{{\mathbf x}}
\def\A{{\mathbf A}}

\def\X{{\mathbf X}}
\def\x{{\mathbf x}}
\def\H{{\mathbf H}}
\def\h{{\mathbf h}}
\def\D{{\mathbf D}}
\def\R{{\mathbb R}}
\def\W{{\mathbf W}}
\def\V{{\mathbf V}}
\def\Z{{\mathbf Z}}

\newtheorem{theorem}{Theorem}
\newtheorem{corollary}{Corollary}
\newtheorem{lemma}{Lemma}
\newtheorem{remark}{Remark}

\begin{document}

\begin{frontmatter}

\title{Pyramidal Reservoir Graph Neural Network}

\author{Filippo Maria Bianchi}
\address{Department of Mathematics and Statistics, UiT the Arctic University of Norway.\\
Hansine Hansens veg 18 -- 9019 Troms{\o} (Norway)}
\ead{filippo.m.bianchi@uit.no}

\author{Claudio Gallicchio\corref{mycorrespondingauthor}}
\address{Department of Computer Science, University of Pisa\\
Largo B. Pontecorvo, 3 -- 57127 Pisa (Italy)}
\ead{gallicch@di.unipi.it}
\cortext[mycorrespondingauthor]{Corresponding author}

\author{Alessio Micheli}
\address{Department of Computer Science, University of Pisa\\
Largo B. Pontecorvo, 3 -- 57127 Pisa (Italy)}
\ead{micheli@di.unipi.it}

\begin{abstract}
We propose a deep Graph Neural Network (GNN) model that alternates two types of layers. The first type is inspired by Reservoir Computing (RC) and generates new vertex features by iterating a non-linear map until it converges to a fixed point. The second type of layer implements graph pooling operations, that gradually reduce the support graph and the vertex features, and further improve the computational efficiency of the RC-based GNN. The architecture is, therefore, pyramidal. In the last layer, the features of the remaining vertices are combined into a single vector, which represents the graph embedding.
Through a mathematical derivation introduced in this paper, we show formally how graph pooling can reduce the computational complexity of the model and speed-up the convergence of the dynamical updates of the vertex features.
Our proposed approach to the design of RC-based GNNs offers an advantageous and principled trade-off between accuracy and complexity, which we extensively demonstrate in experiments on a large set of graph datasets.
\end{abstract}


\end{frontmatter}


\section{Introduction}
\label{sec.intro}

\subsection{Representation learning on graphs with neural networks}

Graphs naturally express entities and their relationships found in many kinds of real data, ranging from molecules to biological and social networks, just to mention a few noteworthy cases. There is an established tradition in processing structured data with Neural Networks, which goes back to the ’90s and recently found a renewed interest in the field of deep learning for graphs~\cite{Wu2020,Bacciu2020Gentle}.
Recursive Neural Networks for processing trees \cite{Sperduti1997,Frasconi1998,hammer_recursive_2004} and directed acyclic graphs \cite{Micheli2004contextual}, provided a neural implementation of a state transition system traversing the input structures to generate the embedding of the input data.   
To process general graphs it was necessary to handle directed cycles and undirected edges, which translate into mutual dependencies among the state variables represented by the neural units.

The  Graph Neural Network (GNN)\cite{scarselli2009graph} and the Neural Network for Graphs (NN4G) \cite{micheli2009neural} started the field following two different lines.
The NN4G pioneered the class of deep spatial graph convolution approaches. In such approaches, the neural units traverse the graphs with shared weights and the mutual dependencies among state variables are modeled by stacking layers, which progressively capture the vertices’ context. 
Recent advancements of the convolutional and related approaches (including kernel based approaches) can be found for instances in  
\cite{defferrard2016convolutional,bianchi2019graph,zhang2018end,tran2018filter,atwood2016diffusion,niepert2016learning,xu2018powerful}. 

The GNN exploits a Recursive Neural Network  to implement a state transition that allows cycles, whereas the contractivity of the state dynamics ensures the stability of the recursive encoding process.
In this approach, the context of each vertex is formed through graph propagation, which is iterated until the dynamics converge to a steady-state. The fixed point of the dynamical system is used  to represent (or {\it embed}) the vertices of input graphs. Theoretical approximation capability and VC dimension of the GNN were recently studied \cite{scarselli2018vapnik}. 

In a context characterized by high complexity and computational cost related to the processing of structured data and the use of deep models,  the need for efficient approaches is becoming more and more important. 
For sequences and trees, the Reservoir Computing (RC) 
\cite{lukovsevivcius2009reservoir}
paradigm provides efficient recurrent/recursive models based on fixed (randomized) values of the recurrent weights under stability conditions of the dynamical system (Echo State Property - ESP) \cite{Jaeger2004,Gallicchio2013tree}.  
A GNN based on RC was introduced in \cite{gallicchio2010graph}, whereas the contractivity of  the reservoir dynamics (ESP) assures the stability condition of the GNN, without any need to alternate training and convergence as for GNN.
The RC model for graphs has been recently  extended to deep architectures in \cite{gallicchio2020fast}, which combines the advantages of a hierarchical abstraction of  the input graphs representation (embedding) through the layers and the efficiency of randomized neural networks.

Another relevant field of graph representation learning is graph pooling, which reduces the dimension of the graph by clustering or by dropping the vertices. 
Global graph representations can be derived by combining at once the embedding of vertex features and use them for inference on downstream tasks at graph level. However, some networks may exhibit scale-dependent behaviors, which have to be accounted for when solving the task. In these cases, pooling operations are exploited to build deep architectures of Neural Networks for graphs, which compute local summaries on the graph to gradually distill the global properties necessary for graph-level inference~\cite{defferrard2016convolutional, bacciu2019non, bianchi2019hierarchical, ying2018hierarchical}.

So far, pooling operations have been only used in conjunction with Neural Networks for graphs that learn vertex representations through functions endowed with parameters optimized with gradient descent~\cite{gilmer2020message}.
This, leaves open the challenge to investigate the contribution of pooling in randomized Neural Networks. 
To fill this gap, in this paper, we propose a further advancement in the design of efficient GNN architecture combing the RC and the graph pooling strategies, exploring their synergy in terms of classification accuracy and computational time.

\subsection{Contributions}
We extend our previous work
in \cite{Bianchi2020pyramidal} and propose a deep architecture composed of RC layers, which generate vertex representations, interleaved with graph pooling operations. 
Since the adopted GNN architecture based on RC generates untrained embeddings, we rely on pooling methods that pre-compute the coarsened graphs without supervision. 
The potential of hierarchical RC architectures in designing fast and deep models for graph classification has been recently shown in \cite{gallicchio2020fast}.
In this context, we investigate the benefit of pooling to further reduce the computational complexity of such randomized neural 
models. 

The contributions of our work are summarized as follows:
\begin{itemize}
    \item We propose a deep pyramidal architecture based on RC for generating graph embeddings. Thanks to suitable graph pooling operations, the proposed model further improves the computational efficiency of RC architectures for graphs.
    \item We derive a mathematical proof of the convergence time in the recursive update necessary to compute the vertex features. This  formal analysis, beside  explaining the speed-up introduced by the pooling operations, links the computational complexity of generic GNNs based on RC with graph theoretical properties of the input data. 
    \item We provide an extensive experimental evaluation of the Pyramidal RC architecture in graph classification tasks, by testing different configurations and considering a large variety of datasets. 
    Notably, we introduce two new datasets for graph classification, with the aim to provide a solid benchmark for testing the proposed method and also other types of GNNs and graph kernels. We also perform dimensionality reduction on the graph embeddings generated by the proposed architecture, which allows us to perform a qualitative evaluation of the embeddings.
\end{itemize}

\subsection{Paper organization}
The remainder of this paper is structured as follows.
We introduce the basic concepts of RC for graphs in Section~\ref{sec.RCgraphs}.
In Section~\ref{sec.pooling} we introduce graph pooling and discuss three pooling operators that pre-compute coarsened versions of the graph based on the topological structure.
Building upon the RC for graphs and graph pooling, we introduce the Pyramidal Reservoir Graph Neural Network architecture in Section~\ref{sec.methodology}.
In Section~\ref{sec.analysis} we perform a theoretical analysis and derive an upperbound of the computational complexity of an RC layer.
The experimental assessment of the approach is given in Section~\ref{sec.experiments}. 
Finally, we draw our conclusions in Section~\ref{sec.conclusions}.

\section{Reservoir Computing for Graphs}
\label{sec.RCgraphs}
In this section we present the elementary concepts of RC for graph structures. We start by introducing the necessary terminology for graph processing used in the rest of the paper in Section~\ref{sec.preliminaries}. Then, in Section~\ref{sec.RGNN}, we describe the RC approach to develop graph embeddings.

\subsection{Preliminaries on Graphs.}
\label{sec.preliminaries}
We denote a graph $\mathcal{G}$ as a tuple $\mathcal{G} = (V_\mathcal{G},E_\mathcal{G})$, where $V_\mathcal{G}$ is the set of vertices and $E_\mathcal{G} \subset V_\mathcal{G}\times V_\mathcal{G}$ is the set of edges. The number of vertices of the graph $\mathcal{G}$ is denoted by $N_\mathcal{G}$, the number of edges is denoted by $M_\mathcal{G}$. The pattern of connectivity among the vertices in $\mathcal{G}$ is represented by the $N_\mathcal{G} \times N_\mathcal{G}$ adjacency matrix $\A_\mathcal{G}$, where $\A_\mathcal{G}(i,j) = 1$ whenever there is an edge between vertex $i$ and vertex $j$, and $\A_\mathcal{G}(i,j) = 0$ otherwise. 

While the computational framework that we describe in this paper is able to deal with both directed and undirected graphs, we here restrict our analysis to the undirected case, which leads to symmetric adjacency matrices $A_\mathcal{G}$. Given a vertex $v \in V_\mathcal{G}$, we define its neighborhood as the set of vertices that are adjacent to $v$, denoted as $\mathcal{N}_\mathcal{G}(v) = \{v' \in V_\mathcal{G} : (v,v')\in E_\mathcal{G}\}$. In this context, the degree of a vertex $v\in V_\mathcal{G}$ is defined as the size of its neighborhood $\mathcal{N}_\mathcal{G}(v)$. 
Also notice that in this case the sum of the degrees of all the vertices equals $2 M_{\mathcal{G}}$.

We also introduce the symmetrically normalized adjacency matrix $\tilde \A_\mathcal{G} = \D_\mathcal{G} ^{-1/2}\A_\mathcal{G} \D_\mathcal{G} ^{-1/2}$, where $\D_\mathcal{G} $ is a diagonal degree matrix, i.e. a diagonal matrix containing the degrees of the vertices $v \in V_\mathcal{G}$.
For our purposes, to each vertex $v\in V_\mathcal{G}$ it is associated an $F$-dimensional feature vector $\x(v) \in \R^F$. Let $\X_\mathcal{G} \in \R^{N_\mathcal{G} \times F}$ be the matrix whose $i$-th row contains the feature vector associated to the $i$-th vertex. We can then represent the graph $\mathcal{G}$ by the couple $(\A_\mathcal{G},\X_\mathcal{G})$. 
In what follows, to ease the notation, we drop the subscript $\mathcal{G}$ whenever the reference to the graph in question is unambiguous or evident from the context.

\subsection{Reservoir Graph Neural Networks}
\label{sec.RGNN}
Reservoir Computing (RC) is a popular paradigm for modeling Recurrent Neural Networks based on untrained (but asymptotically stable) hidden dynamics \cite{lukovsevivcius2009reservoir,gallicchio2017deep}.
The RC approach to process graph structures is based on computing representations, or \emph{graph embeddings}, as fixed points of a dynamical system\footnote{The approach of iterating the operation of a neural architecture until reaching the fixed point of the internal representation has recently been revived under the umbrella of deep equilibrium neural models~\cite{bai2019deep,bai2020multiscale}.}.  The approach has been introduced in \cite{gallicchio2010graph} under the name of Graph Echo State Network, and subsequently extended towards deep architectures in \cite{gallicchio2020fast}, under the name of Fast and Deep Graph Neural Network (FDGNN). 
Crucially, as long as the embeddings are obtained by a stable system, the weights on the internal connections of the neural architecture can be left untrained, and training algorithms can be limited to a simple readout layer. Here we summarize the salient aspects of graph processing by untrained recursive layers in a deep RC-based neural network for graphs, hereafter referred to as \textit{Reservoir Graph Neural Network} (RGNN).

\begin{figure*}
    \centering
    \includegraphics[width=\textwidth]{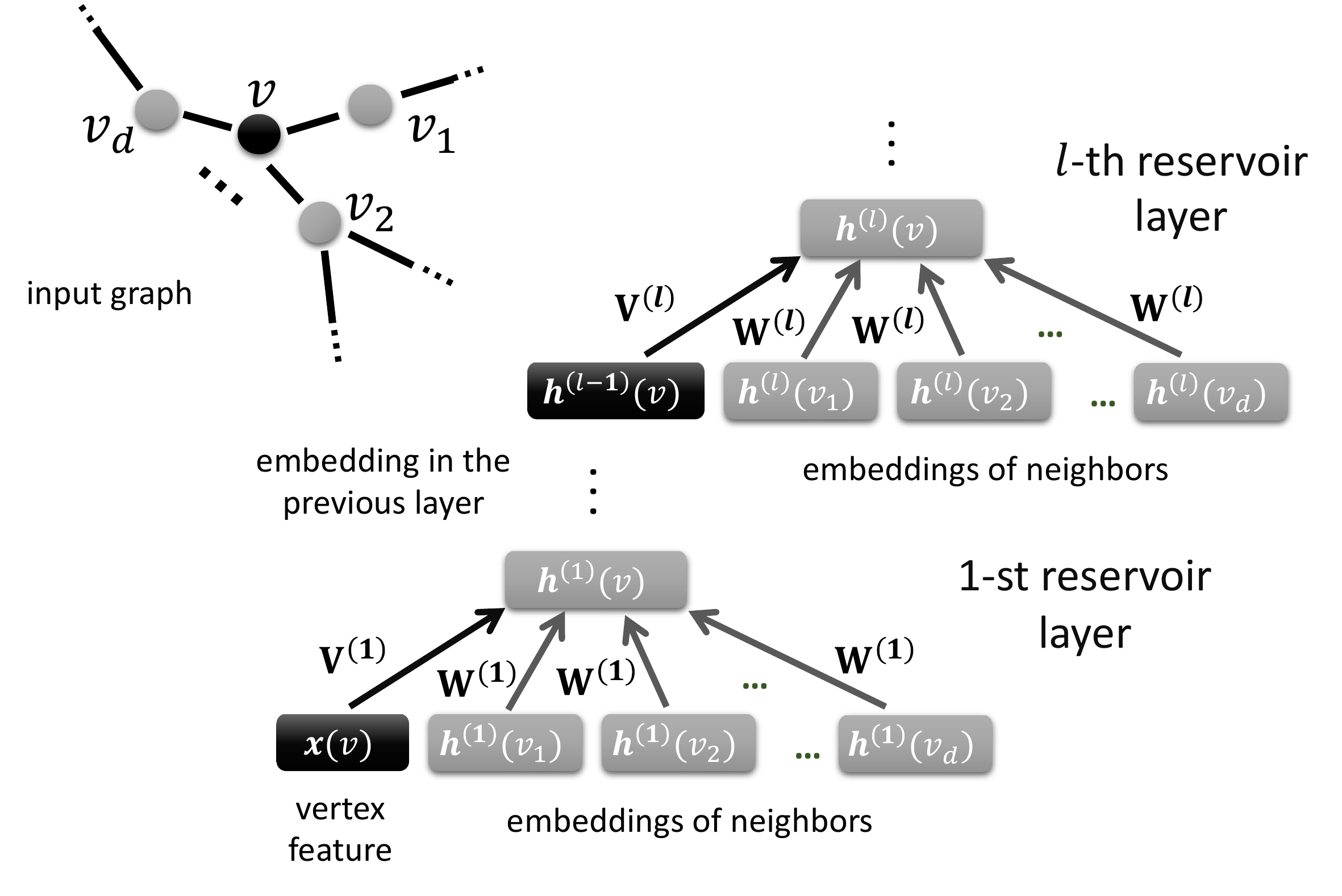}
    \caption{Stack of RGNN layers for graph embedding, applied to a vertex $v$ in the input graph (top left).}
    \label{fig.reservoir}
\end{figure*}

We consider a neural architecture for graphs composed of a number of $L$ hidden recursive reservoir layers, where each layer $l$ computes an embedding for a vertex $v$ by means of the following function:
\begin{equation}
\label{eq.reservoir}
\h^{(l)}(v) = \tanh \left(\sum_{v' \in \mathcal{N}(v)} \h^{(l)}(v') \W^{(l)} + \x^{(l)}(v) \V^{(l)} \right),
\end{equation} 
where $\h^{(l)}(v) \in \R^{H}$ is the embedding, or \emph{state} computed for vertex $v$  by layer $l$. Matrix $\W^{(l)} \in \R^{H \times H}$ is the recurrent weight matrix and modulates the contribution of the neighbors information in determining the embedding for the vertex $v$. Moreover, $\x^{(l)}(v)$ is the feature vector that plays the role of input information for the layer $l$ under consideration. Following the deep architectural construction, $\x^{(1)}(v)$ is the feature vector associated to $v$ in the input graph, i.e. $\x^{(1)}(v) = \x(v)$. For $l>1$, the feature vector $\x^{(l)}(v)$ is instead obtained as the embedding  
computed 
by the previous layer for the same vertex, i.e.  $\x^{(l)}(v) = \h^{(l-1)}(v)$. Accordingly, matrix $\V^{(l)}$ modulates the influence of the input information in determining the embedding, with $\V^{(1)} \in \R^{H \times F}$ for the first layer, and $\V^{(l)} \in \R^{H \times H}$ for layers $l>1$. 
Notice that \eqref{eq.reservoir} is introduced here to set the ground for the mathematical description of the deep RC model, and it is the basic building block of the iterated system on graphs described later in this section (see \eqref{eq.iterated}). Essentially, the states for all vertices (i.e., the $\h^{(l)}(v)$ embeddings) are started with (usually $\mathbf{0}$) initial conditions and then \eqref{eq.reservoir} is iterated for all vertices until a fixed point is reached. The computation then continues with the successive layer until the last one in the architecture is reached.
To ease the architectural setup and description, we are making an assumption\footnote{Using the same number of neurons in every reservoir layer is a simple and commonly adopted choice of architectural design in Deep RC, although the models are generally more flexible and allow a different choice of $H$ for every layer.} here  that every reservoir layer has the same number of neurons $H$. The computation of graph embeddings by reservoir layers in RGNNs is illustrated in Figure~\ref{fig.reservoir}.

For any given layer in the deep architecture, $l = 1, \ldots, L$, the state transition equation in \eqref{eq.reservoir} is applied to every vertex $v$ of an input graph. Instead of considering the operations on the vertices in isolation, it is convenient to represent the embedding operation performed by a RGNN layer on the entire graph. This can be done by grouping in a row-wise fashion the states for all the $N$ vertices of the graph computed at layer $l$ in a matrix $\H^{(l)}\in \R^{N \times H}$, and analogously grouping the feature vectors in a matrix $\X^{(l)}$ (with $\X^{(1)} \in \R^{N \times F}$ for the first layer, and $\X^{(l)} \in \R^{N \times H}$ for $l>1$).
The embedding of the entire graph computed by the $l$-th RGNN layer can then by described as follows:
\begin{equation}
\label{eq.reservoirgraph}
\H^{(l)} = \tanh \left( \A \H^{(l)} \W + \X^{(l)} \V^{(l)} \right),
\end{equation}
which expresses a recursive relation in $\H^{(l)}$. In general, existence and uniqueness of solutions of \eqref{eq.reservoirgraph} is not ensured (for instance, when the input graph contains cyclic substructures or undirected connections). Fortunately, taking the RC approach, we can study \eqref{eq.reservoirgraph} as the function governing a dynamical system and impose a global asymptotic stability property to its dynamics, called Graph Embedding Stability (GES) \cite{gallicchio2020fast}, to guarantee existence and uniqueness of solutions. 
Under the GES property, the system in \eqref{eq.reservoirgraph} will converge to a fixed point (that depends on both $\A$ and $\X$) upon repeated iteration. In \cite{gallicchio2020fast} two conditions have been introduced, one sufficient and one necessary for initializing the $\W^{(l)}$ matrices to satisfy the GES stability property. 

Essentially, to control the stability of the reservoir we need to carefully control the eigenvalues of the system in \eqref{eq.reservoirgraph}, which entails considering also the degree information in the computation (see \cite{gallicchio2020fast} for details). Interestingly, we can use an alternative formulation in which we make use of the normalized adjacency matrix $\tilde \A$ instead of the adjacency matrix $\A$:
\begin{equation}
\label{eq.reservoir2}
\H^{(l)} = \tanh \left( \tilde \A \H^{(l)} \W + \X^{(l)} \V^{(l)} \right).
\end{equation}
Using \eqref{eq.reservoir2} is convenient as it allows us to control the stability of the reservoir system for graphs by simply controlling the eigenvalues of $\W^{(l)}$, as in conventional RC approaches. Accordingly, $\W^{(l)}$ is initialized randomly with values from a uniform distribution in $(-1,1)$, and then re-scaled to have a desired 
\emph{spectral radius}\footnote{I.e., the largest among its eigenvalues in modulus.}, denoted as  $\rho^{(l)}$, exploring values in the range $(0,1)$ for stability \cite{gallicchio2020fast}.
Similarly, the elements in $\V^{(l)}$ are randomly initialized from a uniform distribution on $(-1,1)$, and then re-scaled. For the first layer, we use $\omega_{in}$ as \emph{input scaling}  coefficient, while for successive layers $l>1$ we use $\omega_{hid}^{(l)}$ as \emph{hidden scaling} coefficient (between two consecutive reservoir layers). Usually, the same values of $\rho$ and $\omega_{hid}$ are shared among all the layers. As in common RC settings, $\rho$, $\omega_{in}$ and $\omega_{hid}$ are treated as hyper-parameters of the network (to be fitted on a validation set for each task at hand). 

To compute the embedding for a given graph, we use \eqref{eq.reservoir2} as the iterated map of a discrete-time dynamical system, i.e.:
\begin{equation}
\label{eq.iterated}
\H^{(l)}[t+1] = \tanh\left(\tilde \A\; \H^{(l)}[t]\; \W^{(l)} + \X^{(l)}\, \V^{(l)} \right),
\end{equation}
which is applied until convergence to its fixed point $\H^{(l)*}$. 
As initial condition, it is common to set $\H^{(l)}[0] = \mathbf{0}$ for all layers. 
In practice, for each RGNN layer \eqref{eq.iterated} is iterated until the distance between two consecutive iterations becomes smaller than a convergence threshold $\epsilon$, i.e. $\| \H^{(l)}[t+1] - \H^{(l)}[t]\|_2 < \epsilon$, or until a maximum number of iteration $max_{iter}$ is reached.
The encoding operation proceeds from the first reservoir layer up to the last one in the architecture, bringing to convergence the dynamical systems one layer after the other.
In this regard, notice that for $l>1$, the $\X^{(l)}$ term in \eqref{eq.iterated} corresponds to the fixed point reached by the previous layer in the deep architecture, i.e., $\X^{(l)} = \H^{(l-1)*}$.
\\

\section{Graph pooling}
\label{sec.pooling}

Pooling operators in GNN architectures can be partitioned in two main classes: \textit{feature-based} and \textit{topological} pooling operators.\\

\textbf{Feature-based pooling} methods compute a coarsened version of the graph through differentiable functions, which can be used to learn a cluster assignment matrix that aggregates graph vertexes into super-nodes~\cite{ying2018hierarchical, bianchi2020spectral} or to implement a selection function that determines if the vertices should be kept or dropped~\cite{graphunet, lee2019selfattention}.
The differentiable functions usually take as input the vertex representations, which change across the different layers of the GNN during training. The functions are parametrized by weights that are learned end-to-end, along with the other GNN parameters, by minimizing the loss of a downstream task (e.g., classification or regression loss) and additional (optional) regularization losses. 
Feature-based pooling methods are very flexible, as they adapt the coarsening on the data and the task at hand.
However, GNNs with feature-based pooling layers have more parameters and, thus, their training is slower and more difficult.
Most importantly, there is a weak synergy between feature-based pooling and RGNN layers, since the former require to be trained with gradient descent, while the latter generate unsupervised vertex representation using randomized and untrained weights.\\

\textbf{Topological pooling} methods accounts only for the graph topology described by the adjacency matrix $\A$ and they are typically unsupervised. 
These methods pre-compute the coarsened graphs by optimizing some unsupervised loss function, such as a graph cut or a vertex clustering objective.
The result of the coarsening procedure is, therefore, not influenced by the vertex features or by a downstream task, like in the case of feature-based methods.
Pre-computing graph coarsening not only relieves the need of adapting additional weights with supervised training, but also provides a strong inductive bias that helps to avoid degenerate solutions, such as entire graphs collapsing into a single node or entire graph sections being discarded.
This is important when dealing with small datasets in tasks such as graph signal classification.
Topological pooling methods can be effectively combined with reservoir computing layers, which fit their vertex representations to the pre-computed coarsened graphs.
By alternating reservoir computing and topological pooling layers, it is possible to gradually extract hierarchical graph features in a completely unsupervised fashion. 

In the following, we provide the details of the three different topological pooling methods that are considered in this paper. 
First, in Section~\ref{sec:graclus} we introduce \textit{Graclus} pooling. 
Then, in Section~\ref{sec:nmf} we illustrate the \textit{Non-negative Matrix Factorization} pooling. 
Finally, in Section~\ref{sec:ndp} we describe the recently introduced \textit{Node Decimation Pooling}.

\subsection{Graclus pooling}
\label{sec:graclus}
\textit{Graclus} is a hierarchical spectral clustering algorithm that has been originally proposed in \cite{dhillon2004kernel} and it can be used to compute coarsened graphs.
\textit{Graclus} has been implemented in several GNN architectures to perform pooling~\cite{bruna2013spectral, defferrard2016convolutional, fey2018splinecnn, levie2017cayleynets}.
The $l$-th pooling operation first clusters together two vertices $v_i$ and $v_j$ into a new vertex $v_k$.
Then, a standard pooling operation (average or max pool) is applied to compute the vertex feature $\x^{(l+1)}(v_k)$ of the new vertex $v_k$ from $\x^{(l)}(v_i)$ and $\x^{(l)}(v_j)$.
To make the pooling output consistent with the clusters assignment, the rows of graph signal $\X$ are rearranged so that vertices $v_i$ and $v_j$ end up in consecutive positions.

\textit{Graclus} pooling has few drawbacks.
First, the connectivity of the original graph is not preserved in the coarsened graphs and the spectrum of their associated Laplacians might not be contained in the spectrum of the original Laplacian.
Second, \textit{Graclus} pooling adds ``fake'' vertices by constructing a balanced binary tree, so that the number of vertices can be exactly halved at each pooling step.
Unfortunately, this not only injects noisy information in the graph, but also increases the computational complexity in the GNN, both in the training and in the evaluation step.

\subsection{Non-negative Matrix Factorization (NMF) pooling}
\label{sec:nmf}
Originally proposed in 
\cite{bacciu2019non}, this approach exploits the non-negative matrix factorization to cluster the rows (or the columns) of the adjacency matrix of the graph.
Such an approach is slightly different from \textit{Graclus}, where the vertices of the graph are partitioned with spectral clustering.
In particular, the pooling algorithm uses the \textit{NMF} factorization such that 
$\mathbf{A} \approx \mathbf{Q}\mathbf{S}$, where $\mathbf{Q} \in \mathbb{R}^{N \times K}$ has the role of cluster representatives matrix, and $\mathbf{S} \in \mathbb{R}^{K \times N}$ gives a soft-clustering of the columns in $\mathbf{A}$. 
Given the vertices features before the pooling $\X$, the new vertices features are computed as $\mathbf{X}_\text{pool} = \mathbf{S}^T\mathbf{X}$. The new adjacency matrix is computed as $\mathbf{A}_\text{pool} = \mathbf{S}^T \mathbf{A} \mathbf{S}$.

The main drawback of this pooling method is that \textit{NMF} does not scale well to large graphs.
Indeed, both the space and memory complexity are cubic in the number of vertices, meaning that \textit{NMF} cannot be computed only for relatively small graphs~\cite{vavasis2010complexity}.
A second drawback is that the coarsened graphs obtained with \textit{NMF} pooling are extremely dense.
This poses another scalability issue, since dense graphs are expensive to store and the sparse operations (commonly used to implement multiplications with the adjacency matrix) become extremely inefficient if the graph density is too high.

\subsection{Node Decimation Pooling (NDP)}
\label{sec:ndp}
This pooling scheme for GNNs, originally proposed in 
\cite{bianchi2019hierarchical}, computes coarsened graphs in three steps.
\begin{enumerate}
    \item The vertices of the original graph are divided in two sets according to the MAXCUT partition and then one of the two sides of the partition is dropped, reducing the number of vertices approximately by half. If two vertices are strongly connected on the graph, i.e., there are many paths that connects them, they will be assigned to opposite sides of the MAXCUT partition.
    The idea is that such strongly connected vertices will exchange a lot of information during the message-passing operations implemented by the GNN and, therefore, their content will be very similar and redundant. In this sense, one side of the MAXCUT partition will summarize well the whole graph.
    \item After dropping the vertices and all their incident edges, the resulting graph is likely to be disconnected.
    The remaining vertices are connected together by a link construction procedure based on Kron reduction~\cite{kron_red}.
    Kron reduction enjoys the connectivity preservation property, i.e., it generates a new graph where two vertices are connected if and only if there was a path between them in the original graph. Additionally, Kron reduction does not introduce self-loops, guarantees the preservation of resistance distance, and preserves the spectrum of the original graph~\cite{shuman2016multiscale}. As we will discuss later, these properties prevent the spectrum of the coarsened graph to grow and allows a faster convergence of the RGNN layers.
    \item A sparsification procedure allows to prune the \textit{weak} edges from the coarsened graph, i.e., those edges associated with small weights in the weighted adjacency matrix. The sparsification procedure is controlled by an hyper-parameter $\delta$ that determines the amount of edges to be dropped. The graph sparsification allows to drop a large amount of edges at the cost of compromising the graph structure. In this sense, graph sparsification trades the computational cost (which scales linearly with the number of edges) with the quality of the graph representations
\end{enumerate}

Each method gives in output a coarsened graph, $\mathbf{A}_\text{pool}$ and a pooling matrix $\mathbf{S}$ used to compute the new vertex features,  $\mathbf{X}_\text{pool} = \mathbf{S}^T\mathbf{X}$.
In the case of \textit{Graclus} and \textit{NMF}, $\mathbf{S}$ is a matrix that defines how the vertices should be clustered together.
Instead, in \textit{NDP} $\mathbf{S}$ is a selection matrix that identifies the vertices that should be kept or dropped.

\section{Pyramidal Reservoir GNN}
\label{sec.methodology}

In this section we describe the proposed modular GNN architecture composed of $L$ blocks, where RGNN layers are interleaved by pooling layers.

A pooling operator can be applied recursively to compute increasingly coarsened versions of the adjacency matrix.
We introduce the notation $p^{l}(\A)$ to indicate that the pooling operator has been applied $l$ times to $\A$.
For example:
\begin{itemize}
    \item $\A^{(1)} = p^0(\A) = \A$,
    \item $\A^{(2)} = p(\A) = \A_\text{pool}$,
    \item $\A^{(3)} = p^2(\A) = p(p(\A)) = p(\A_\text{pool})$.
\end{itemize}

The $l$-th block consists of an RGNN layer and a pooling layer.
The RGNN layer computes the new vertex features $\H^{(l)}$, using the vertex features $\X^{(l)}$ and the adjacency matrix $\A^{(l)}$.
The pooling layer generates a pooled version of the vertex 
embeddings
$\X^{(l+1)} = \left(\mathbf{S}^{(l)}\right)^T\H^{(l)}$, which become the input vertex features for the RGNN layer in the next block.
The last block in the architecture consists only of a RGNN layer.
The pooling matrices $\mathbf{S}^{(l)}, l=1, \dots, L-1$ and the coarsened adjacency matrices $\A^{(l)} = p^{l-1}(\A), l=1, \dots, L$, are obtained according to one of the three graph pooling scheme, \emph{Graclus}, \emph{NMF} or \emph{NDP}, described in Section~\ref{sec.pooling}.

\begin{figure*}[!ht]
    \centering
    \includegraphics[width=\textwidth]{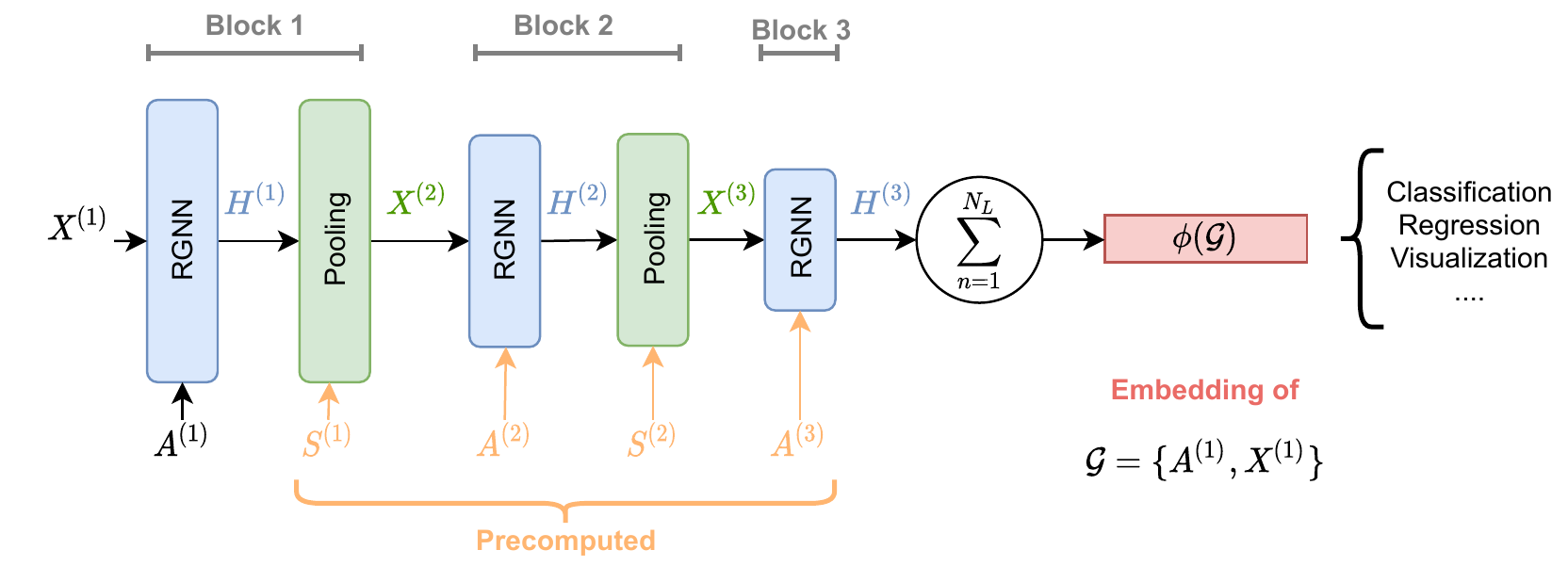}
    \caption{
    Schematic depiction of a Pyramidal RGNN with $L=3$ blocks. 
    In every block $l$, the RGNN layer computes the new vertex features $\H^{(l)}$ from the vertex features in the original input graph, $\X^{(1)}$ (or from those generated by the previous block $\X^{(l)}$, for $l>1$), and from the $l$-th version of the adjacency matrix, $\A^{(l)}$. 
    The pooling layer reduces the vertex features, according to a pooling matrix $\mathbf{S}^{(l)}$. 
    The corasened versions of the adjacency matrix $\A^{(l)}, l=1, \dots, L$, and the pooling matrices $\mathbf{S}^{(l)}$, $l=1,\dots, L-1$ are pre-computed by a graph pooling algorithm.
    The vertex features of the last layer, $\H^{(L)}$, are aggregated all together to form the graph embedding $\phi(\mathcal{G})$, which can be used to perform different types of supervised or unsupervised tasks.}
    \label{fig:pyramid_rgnn}
\end{figure*}

After the last pooling operation, the number of remaining vertices $N_L$ is usually less than the number of vertices in the original graph, but greater than 1, i.e., $1 \leq N_L \leq N$.
Therefore, to obtain a vectorial graph representation, we combine the $N_L$ vertex features with an aggregation operator.
Common aggregation operators are the average and the sum; we used the latter in this work and obtained a graph embedding $\phi(\mathcal{G}) \in \mathbb{R}^H$, defined as
\begin{equation}
    \label{eq:graph_embedding}
    \phi(\mathcal{G}) = \sum_{n=1}^{N_L} \H^{(L)}[n,:],
\end{equation}
where $\H^{(L)}[n,:]$ is the $n$-th row of the matrix of vertex features in the last block $L$ of the architecture.
The sum aggregation can be considered as a global pooling operation, which replaces the local pooling layer in the last block.
Figure~\ref{fig:pyramid_rgnn} reports a schematic depiction of a Pyramidal RGNN with $L=3$ blocks: [\texttt{RGNN}, \texttt{Pool}]-[\texttt{RGNN}, \texttt{Pool}]-[\texttt{RGNN}].

Both the RGNN and the topological pooling layers do not require any form of supervised training: the first exploits randomized weights to compute the vertex representations in a stable fashion, while the latter pre-computes coarsened graphs by optimizing some unsupervised loss, such as a clustering objective.
Therefore, each graph embedding $\phi(\mathcal{G})$ is an \textit{unsupervised} vectorial representation of the original graph $\mathcal{G} = (\A_\mathcal{G},\X_\mathcal{G})$.

The vectorial representations can be used to solve supervised downstream tasks, such as graph classification or graph regression, by means of common classification or regression models for vectorial data that act as readout layer.
In line with typical RC approaches \cite{lukovsevivcius2009reservoir}, in this paper we use a linear readout layer trained by ridge regression when we use the graph embeddings $\phi(\mathcal{G})$ to solve classification tasks.
Using a linear readout is also aligned with common approaches followed by other types of randomized architectures~\cite{scardapane2017randomness}.
We notice that since the graph embeddings $\phi(\mathcal{G})$ are computed without supervision, they can, in principle, also be used in tasks such as clustering and unsupervised dimensionality reduction.

\section{Convergence Analysis and Computational Cost}
\label{sec.analysis}

In this section, we analyze the cost of running RGNNs on graph structures. The aim is to show the connection between this cost and some major structural properties of the input graph in question (number of vertices, number of edges, and spectrum of the graph). First, in Section~\ref{sec.convergence} we study the convergence behavior of RGNN layers on graphs.
Then, in Section~\ref{sec.cost} we present the computational cost analysis. 
Notice that Section~\ref{sec.convergence} is mainly devoted to building the mathematical tools necessary for characterizing the asymptotic behavior of contractive RGNN systems. The readers interested in the final results on the computational cost analysis can directly jump to Section~\ref{sec.cost}.

\subsection{Convergence Analysis}
\label{sec.convergence}
In this section we analyze the convergence behavior of reservoir systems on graph structures.
We recall from Section~\ref{sec.RCgraphs} that an RGNN layer with $H$ recursive neurons implements a hidden state dynamics given by the equation\footnote{In this section we remove the layer superscript to ease the notation.}:
\begin{equation}
\label{eq.iterateda}
\H[t+1] = \tanh(\tilde \A\; \H[t]\; \W + \X\, \V),
\end{equation}
where $\H[t]$ is the matrix of graph embeddings, $\tilde \A$ is the normalized adjacency matrix of the graph, $\X$ is the input feature matrix at the current layer, while $\W$ and $\V$ are respectively the recurrent and input weight matrices. Finally, $\tanh$ denotes the element-wise applied non-linearity of the state transition. In what follows, we use $\|\cdot\|$ to refer the 2-norm of a matrix (i.e., its largest singular value).
Equation \eqref{eq.iterateda} is iterated until convergence to its fixed point, with a convergence threshold denoted as $\epsilon$.
Our main result in this section is to link the number of iterations required for convergence, denoted by $T$, to the spectrum of the input graph. We restrict our analysis to the cases in which the state transition function in \eqref{eq.iterateda} is contractive.
In Lemma~\ref{th.lemma} we show that the contraction coefficient for reservoirs on graphs can be expressed as $K = \rho(\tilde \A) \|\W\|$. This result is then used in Theorem~\ref{th.theorem} to 
compute the value of $T$.
Interestingly, we finally observe in Corollary~\ref{th.corollary} that the derived expression for $T$ is an increasing function with $\rho(\tilde \A)$, i.e. the spectral radius of $\tilde \A$.

\begin{lemma}[Lipschitz continuity of reservoir systems on graphs]
\label{th.lemma} 
Consider a reservoir system on graphs governed by the state transition equation in \eqref{eq.iterateda}. The system is Lipschitz continuous with constant $K = \rho(\tilde \A)\|\W\|$. 
\end{lemma}
\begin{proof}
Let us consider one iteration step of the reservoir system in \eqref{eq.iterateda}, applied to the graph $(\tilde \A,\X)$, and starting from states $\H$ and $\Z$.
Under the action of the reservoir system, states $\H$ and $\Z$ are mapped respectively to $\tanh(\tilde \A \H \W + \X \V)$ and $\tanh(\tilde \A \Z \W + \X \V)$.
The distance between the mapped states can then be bounded as follows:
\begin{equation}
\label{eq.lemma}
\begin{array}{l} 
\|\tanh(\tilde \A \H \W + \X \V) - \tanh(\tilde \A \Z \W + \X \V)\| \leq
\\
\|\tilde \A \H \W + \X \V - \tilde \A \Z \W - \X \V\| =
\\
\| \tilde \A \H \W - \tilde \A \Z \W\| \leq 
\\
\|\tilde \A\| \|\W\| \; \|\H - \Z\| = 
\\
\rho(\tilde \A) \|\W\| \; \|\H - \Z\|,
\end{array}
\end{equation}
where in the last step we have used the symmetry of the adjacency matrix  ($\|\tilde \A\| = \rho(\tilde \A)$).
From the last line in the above \eqref{eq.lemma} it is straightforward to observe that $K = \rho(\tilde \A) \|\W\|$ is a Lipschitz constant of the reservoir system.
\end{proof}
In light of the result of Lemma~\ref{th.lemma}, we observe that whenever the reservoir system satisfies the condition $K = \rho(\tilde \A) \|\W\| < 1$ the resulting iterated map is a contraction and the system globally converges to its unique fixed point independently on the initial conditions. This observation is in line with the sufficient condition for the GES property in \cite{gallicchio2020fast}. In this case, we can derive an 
expression
for $T$, the number of iterations required for convergence.

\begin{theorem}[Convergence of reservoir systems on graphs]
\label{th.theorem}
Consider a contractive reservoir system on graphs governed by the state transition equation in \eqref{eq.iterateda}. Given a convergence threshold $\epsilon$, the system approaches the fixed point within a number of iterations $T$ given by:
\begin{equation}
\label{eq.theorem1}
T = \left\lceil{  \frac{\ln \epsilon + \ln (1-\rho(\tilde \A)\|\W\|) - \ln H_1}{\ln \rho(\tilde \A)\|\W\|} }\right\rceil,
\end{equation}
where 
$H_1 = \|\tanh(\X \V)\|$ is the norm of the reservoir state after the first iteration.
\end{theorem}
\begin{proof}
Let us denote by $\H_0$, $\H_1$, $\H_t$, and $\H^*$ respectively the initial conditions, the first, the $t$-th iterate and the fixed point of equation~\eqref{eq.iterateda}. Let $K$ denote a contraction coefficient of the reservoir system. As $\H^*$ is not modified by \eqref{eq.iterateda}, it is straightforward to see that 
\begin{equation}
\label{eq.th1}
\|\H^*-\H_t\|\leq K^t \|\H^*-\H_0\|.
\end{equation}
We also observe that $\|\H^*-\H_0\| \leq \| \H^*-\H_1 \| + \| \H_1-\H_0 \|$, from which it follows that $\| \H^*-\H_0 \| \leq K \| \H^*-\H_0 \| + \| \H_1-\H_0 \|$, hence:
\begin{equation}
\label{eq.th2}
\| \H^*-\H_0 \| \leq \frac{1}{1-K} \|\H_1-\H_0 \|.
\end{equation}
Now, using \eqref{eq.th2} into \eqref{eq.th1} we have that:
\begin{equation}
\label{eq.th3}
\|\H^* - \H_t\| \leq \frac{K^t}{1-K}\|\H_1 - \H_0\|,
\end{equation}
which is interesting as it allows us to estimate the distance between the fixed point and any $t$-th iterate by only knowing the contraction coefficient and the distance between the first iterate and the initial conditions. 
Successive iterations will then bring the system closer and closer to the fixed point, until the distance becomes smaller than the convergence threshold $\epsilon$. This certainly occurs for $\frac{K^t}{1-K}\|\H_1 - \H_0\| \leq \epsilon$, which leads to $K^t \leq \frac{\epsilon (1-K)}{\|\H_1-\H_0\|}$, and (observing that $K<1$)  to $t \geq \log_K{\frac{\epsilon (1-K)}{\|\H_1-\H_0\|}}$.
A number of iterations after which we are sure to be not farther than $\epsilon$ from the fixed point is then $T = \left\lceil{ \log_K{\frac{\epsilon (1-K)}{\|\H_1-\H_0\|}} }\right\rceil$.
Using a zero initial condition $\H_0 = \mathbf{0}$, the Lipschitz constant resulting from Lemma~\ref{th.lemma}, and expanding the logarithm, we finally derive that:
\begin{equation}
\label{eq.thlast}
T = \left\lceil{\frac{\ln \epsilon + \ln (1-\rho(\tilde \A)\|\W\|)-\ln H_1}{\ln (\rho(\tilde \A)\|\W\|)}}\right\rceil.
\end{equation}

\end{proof}

Theorem~\ref{th.theorem} tells us that the number of iterations required for convergence depends on several factors. Besides the dependence on the desired convergence tolerance $\epsilon$, some of these factors are related to the RGNN hyper-parametrization ($\W$ and $\V$), others depend on the input graph features ($\X$) and structure ($\tilde \A$). Interestingly, keeping fixed the reservoir weight matrices and the input features, the number of iterations $T$ is smaller for smaller values of $\rho(\tilde \A)$, as expressed by the following Corollary~\ref{th.corollary}.

\begin{corollary}
\label{th.corollary}
Consider a contractive RGNN layer on graphs governed by the state transition equation ~\eqref{eq.iterateda}. Keeping fixed the RGNN weight matrices and the graph input features, the number of iterations required for convergence is an increasing function with $\rho(\tilde \A)$.
\begin{proof}
This result follows straightforwardly from Theorem \ref{th.theorem}. For instance, note that 
$\epsilon (1-\rho(\tilde \A)\|\W\|) / H_1$ is a decreasing function of $\rho(\tilde \A)$, and that
$\log_K(\cdot)$, with $K<1$, is a decreasing function of its argument. Composing two decreasing functions gives an increasing function, hence $T$ is increasing with $\rho(\tilde \A)$.
\end{proof}
\end{corollary}
Hence, we can express $T$ as $\mathcal{T}(\rho(\tilde \A))$, an increasing function of $\rho(\tilde \A)$.

\subsection{Computational Complexity of RGNN}
\label{sec.cost}

We analyze here the computational cost of running an RGNN layer on graphs, recalling that for each input graph the embedding computation requires to iterate \eqref{eq.iterateda} until convergence to its fixed point.
Noticing that the matrix product $\X\, \V$ in \eqref{eq.iterateda} can be performed only once, the cost per each iteration scales as $\mathcal{O}( M\,H+N\,H^2)$, where 
$N$ and $M$ are respectively the number of vertices and the number of edges in the graph, and $H$ is the number of reservoir neurons in the layer\footnote{The cost can be made linear in $H$ by using sparse recurrent weight matrices $\W$. However, in this paper we focus the analysis on the input graph structure, rather than trying to optimize the reservoir matrix operations.}.
The overall cost 
per layer is $\mathcal{O}(T\,( M\,H+N\,H^2)\,)$, with $T$ the number of iterations required for convergence. As shown in the previous Section~\ref{sec.convergence} (see Corollary~\ref{th.corollary}), under the assumption of contractive reservoir dynamics\footnote{Although this assumption could be too strong to hold always in practice (like in the case of convetional RC neural networks \cite{lukovsevivcius2009reservoir}), the derived insights can be of interest in general to guide the network design.}, $T$ can be easily expressed as an increasing function of $\rho(\tilde \A)$, i.e. $\mathcal{T}(\rho(\tilde \A))$.
Thereby, the computational complexity for graph embedding computation by RGNN can be finally expressed as:
\begin{equation}
\label{eq.cost}
\mathcal{O}\left( 
\mathcal{T}\big(\rho(\tilde \A)\big)
\big( M\,H+N\,H^2\big)
\right).
\end{equation}

From the above \eqref{eq.cost} we can derive a number of interesting remarks. 
\begin{remark}
\label{remark.1}
Relevantly for the scopes of this paper,
we see that, keeping fixed the reservoir weight matrices, the cost 
of the embedding computation
increases with the number of vertices ($N$), with the number of edges ($M$), and with the largest absolute eigenvalue of the normalized adjacency matrix ($\rho(\tilde \A)$).
This allows us to motivate in a mathematically grounded way the use of pooling operators interleaved between successive layers of RGNN. In fact, while trying to preserve as much as possible the information content of the original graph, pooling can act on all 3 indicated  factors ($N$, $M$, and $\rho(\tilde \A)$), with a substantial impact on the resulting computation time.
\end{remark}

\begin{remark}
The cost of the embedding is the same for both training and test, as the reservoir parameters do not undergo any training process. This also implies that \eqref{eq.cost} gives a lower bound to the cost of the embedding process for the entire class of recursive GNNs (possibly with training).
\end{remark}

\begin{remark}
Whenever the set of graphs of interest has a bounded degree - a condition that is typical in many application domains such as chemistry - the cost scales \emph{linearly} with the number of vertices.
\end{remark}

\begin{remark}
In deep architectural settings, when the neural network comprises a composition of multiple RGNN layers, the cost in \eqref{eq.cost} needs to be accounted for each layer.
This implies a straightforward advantage of the deep architectural settings versus a shallow one 
for the same total number of recursive hidden units, in line with the cases of deep RC for time-series \cite{gallicchio2018design} and trees \cite{Gallicchio2018deeptree}.
\end{remark}

\section{Experiments}
\label{sec.experiments}
In our recent work, 
we showed the competitive performance of RGNN-based architectures in terms of classification accuracy~\cite{gallicchio2020fast, gallicchio2020ring}.
By using 
a sufficiently deep RGNN  
it is possible to achieve
a classification accuracy that is on par or even superior to state-of-the-art neural networks and kernel for graphs \cite{gallicchio2020fast}.

Here, instead, we focus on studying the trade-off between classification accuracy and computing time when a deep RGNN is equipped with graph pooling operations.
We perform the experimental evaluation by using an architecture that is simpler compared to 
that in \cite{gallicchio2020fast},
as we use 
fewer layers and hidden units, as well as a linear readout.
A simpler model allows to better understand the effect of the pooling operations and to relate our results to the theoretical analysis presented in Section~\ref{sec.analysis}.

In our experiments, we consider two groups of datasets for graph classification.
The first group comprises two new datasets proposed for the first time in this paper and is described in Section~\ref{sec:new_dataset}.
Then, as a second group, we consider some popular benchmark datasets for graph classification, summarized in Section~\ref{sec.tud}.
We perform two types of experiments.
First, in Section~\ref{sec.classification} we evaluate the performance, in terms of classification accuracy and computing time, achieved by different pooling approaches.
Then, in Section~\ref{sec.visualization} we perform a qualitative analysis of the graph embeddings by visualizing them with a supervised dimensionality reduction technique.

All the experiments were performed on a server with an Intel i7-10700K processor, two Nvidia RTX-2080Ti, and 64GB of RAM.
A library that allows to implement the proposed Pyramidal RGNN is publicly available online\footnote{\url{https://github.com/FilippoMB/Pyramidal-Reservoir-Graph-Nerual-Networks}}.

\subsection{New benchmark datasets for graph classification}
\label{sec:new_dataset}
An important shortcoming in several benchmark dataset for graph classification is that in some cases the vertex features contain enough information to achieve high classification accuracy, even without considering the graph structure.
Similarly, in other datasets the vertex features are trivial or they are not present at all.
This problem has been highlighted by recent works, which showed that very simple baselines that disregard either the graph connectivity or the vertex features are able to achieve similar performance to elaborated state-of-the-art models~\cite{orlova2015graph, errica2019fair, bianchi2020spectral}.

\begin{figure*}
    \centering
    \includegraphics[width=\textwidth]{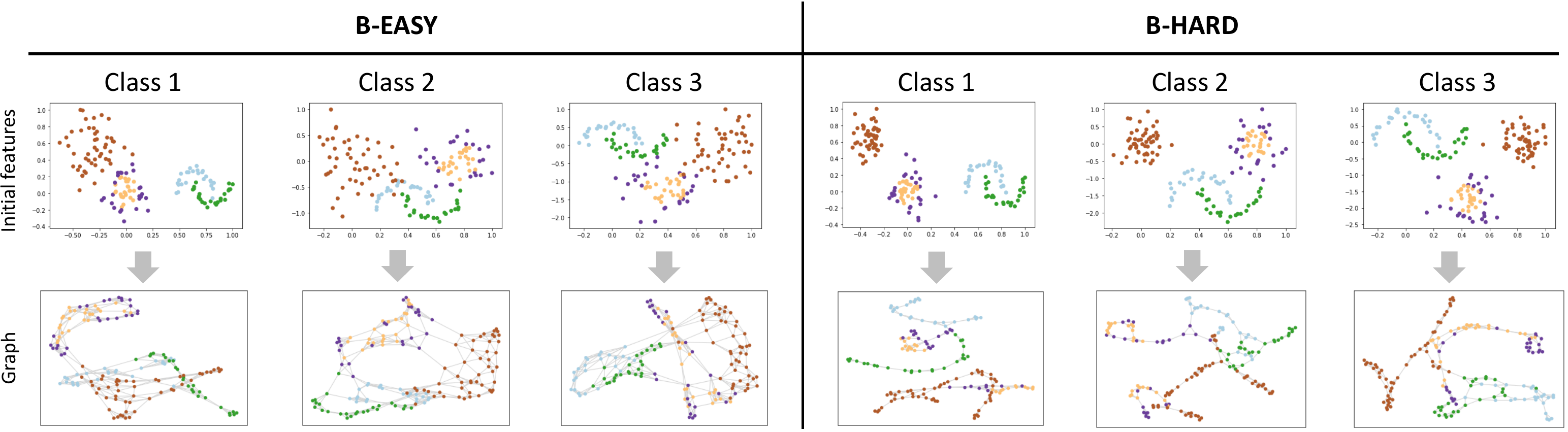}
    \caption{The figure depicts an example of how the graphs of the 3 different classes are generated in the two benchmark datasets for graph classification. Each class is determined by the relative positions of the clusters of 2-D points. 
    For example, in class 1 there is a Gaussian blob on the left, concentric circles in the middle, and two-moons on the right.
    In the \textit{easy} version the clusters are more scattered and overlapping, while in the \textit{hard} version are more compact.
    A $k$-NN graph with $k=5$ for the version \textit{easy} and $k=4$ for \textit{hard} is used to generate a graph from the original data points.}
    \label{fig:bench_datasets}
\end{figure*}

Motivated by these reasons, we propose 
a new class of datasets
where both the vertex features and the adjacency matrix are completely uninformative if considered alone. Therefore, an algorithm that relies only on the vertex features or on the graph structure will fail to perform better than a random classifier.
The proposed datasets consist
of graphs belonging to 3 different classes. 
Each graph is constructed by first generating 5 random clusters of points, each one with a particular shape (Gaussian blob, two moons, and concentric circles).
The points in each cluster are assigned with a different color, encoded by a one-hot vector in $\mathbb{R}^5$. 
Such vectors are the input features associated with each vertex and represent the rows of the vertex feature matrix $\X \in \mathbb{R}^{N \times 5}$. 
The graph is built by connecting the points with a $k$-nn graph and the vertex features are the $\mathbb{R}^5$ one-hot color vectors.
The three classes of graph are determined by the relative position of the original clusters.
An example is depicted in Figure~\ref{fig:bench_datasets}, where the first row shows the original clusters and the second row the resulting graphs, obtained by connecting each point with its closest $k$ neighbors.

Based on this approach we introduce two dataset versions:
\textit{easy} and \textit{hard}.
The graphs in the first version are easier to classify 
than those
in the second one.
The difficulty is controlled by the compactness of the original clusters (compact clusters in the \textit{easy}, higher variance in the cluster of \textit{hard}) and the number of edges in the $k$-nn graph ($k=5$ in \textit{easy} and $k=4$ in \textit{hard}).

The two versions of the dataset are available online\footnote{\url{https://github.com/FilippoMB/Benchmark_dataset_for_graph_classification}} and they are already split in train, validation and test.
The online repository contains also a small version of the two datasets (\textit{easy-small} and \textit{hard-small}), which can be useful during debug and preliminary testing stages.

\subsection{TUD datasets}
\label{sec.tud}

Next, we consider the popular TUD datasets for graph classification~\cite{KKMMN2016}.
We selected a rich variety of datasets, where graphs are either representing molecules (Mutagenicity, Proteins, Enzymes, NCI1, D\&D) or social networks (IMDB-BINARY, IMDB-MULTI, COLLAB, Reddit-2K, Reddit-5K). 

The datasets are very different from each other, in terms of number of graph classes, number of vertices and edges in each graph, and type of vertex attributes and vertex labels.
Therefore, the selected datasets allow to test the performance of the graph classification algorithms in a large variety of use-cases.

Table~\ref{tab:graph_dataset} reports the statistics of each dataset used in our study.
In each dataset, the vertices can be associated  both with a 
``vertex attribute'' 
(a real-valued vector) and a 
``vertex label'' 
(a discrete value represented by a one-hot vector).
Vertex labels 
should not be confused with the graph label, which defines the class
of the entire graph.
The node attribute and the node label are concatenated to obtain the vertex feature vector $\mathbf{x}$.
We note that in some 
datasets there are neither 
vertex attributes
nor 
vertex labels:
in those cases we use the node degree as surrogate vertex feature.

\bgroup
\def\arraystretch{1} 
\setlength\tabcolsep{.3em} 
\begin{table*}[!ht]
\footnotesize
\centering
\caption{Summary of statistics of the graph classification datasets.} 
\label{tab:graph_dataset}
\begin{tabular}{lcccccc}
\cmidrule[1.5pt]{1-7}
\textbf{Dataset} & \textbf{samples} & \textbf{classes} & \textbf{avg. vertices} & \textbf{avg. edges} & \textbf{vertex attr.} & \textbf{vertex labels} \\
\cmidrule[.5pt]{1-7}
B-easy   & 1,800  & 3  & 147.82 & 922.66   & -- & yes \\
B-hard   & 1,800  & 3  & 148.32 & 572.32   & -- & yes \\
\cmidrule[.5pt]{1-7}
Mutagenicity & 4,337  & 2  & 30.32  & 91.87    & -- & yes \\
Proteins     & 1,113  & 2  & 39.06  & 72.82    & 1  & no  \\
Enzymes      & 600    & 6  & 32.63  & 62.14    & 18 & yes \\
NCI1         & 4,110  & 2  & 29.87  & 32.30    & -- & yes \\
D\&D         & 1,178  & 2  & 284.32 & 715.66   & -- & yes \\
\cmidrule[.5pt]{1-7}
IMDB-BIN     & 1,000  & 2  & 19.77  & 96.53    & -- & no  \\
IMDB-MUL     & 1,500  & 3  & 13.00  & 65.94    & -- & no  \\
COLLAB       & 5,000  & 3  & 74.49  & 2,457.78 & -- & no  \\
Reddit-2K    & 2,000  & 2  & 429.63 & 497.75   & -- & no  \\
Reddit-5K    & 4,999  & 5  & 508,52 & 594,87   & -- & no  \\
\cmidrule[1.5pt]{1-7}
\end{tabular}
\end{table*}
\egroup

\subsection{Graph classification}
\label{sec.classification}

We evaluate the performance in terms of classification accuracy and computational time of the 12 datasets described in Section~\ref{sec:new_dataset} and \ref{sec.tud}.
To generate the graph embeddings, we consider two types of architectures.
The first is the Pyramidal RGNN presented in Section~\ref{sec.methodology}, which is equipped with one of the three pooling operators described in Section~\ref{sec.pooling}: \textit{Graclus}, \textit{NMF} and \textit{NDP}.
The second architecture is identical to the previous one but without pooling layers and is referred to as ``\emph{No-pool}''. In our experimental analysis, we use the \emph{No-pool} model as a baseline reference for comparing the results obtained by the Pyramidal architectures, as this allows us to assess directly the effect of introducing the pooling operations.
For both variants, we considered an architecture both with $L=2$ and with $L=3$ blocks, for a total of 8 architectures.
We also notice that the \textit{No-pool} architecture is similar to the one presented in our previous work on FDGNN~\cite{gallicchio2020fast}, to which we refer the interested reader for a comparison, in terms of classification accuracy, with other state-of-the-art models.

To perform the classification, once the graph embeddings are computed they are processed by a linear ridge regression classifier that is controlled through a Tikhonov regularization hyper-parameter $\alpha$.
The model selection is performed by following the same cross-validation scheme described in \cite{errica2019fair}, with the following setting:
\begin{enumerate}
    \item 5 external folds (test split);
    \item internal hold out (90\% train, 10\% validation);
    \item 100 different random hyper-parameters configurations obtained by uniformly and randomly drawing $\rho \in [.1, .9]$, $\omega_{in} \in [.1, .8]$, $\omega_{hid} \in [.1, .8]$;
    \item For each hyper-parameter configuration, we perform a grid search on the regularization hyper-parameter $\alpha \in \{1e2, 1e1, 1e0, 1e-1, 1e-2\}$;
    \item For each configuration $\{ \rho, \omega_{in}, \omega_{hid}, \alpha \}$ we generate 3 random initialization of the reservoir weights and compute the average accuracy on the validation set;
    \item The model defined by the optimal configuration $\{\rho^{*}, \omega_{in}^{*}, \omega_{hid}^{*}, \alpha^{*}\}$, i.e., the one that achieves the highest average accuracy on the validation set, is tested on the external fold (test set);
    \item The average accuracy obtained on all the 5 (test) folds is reported as the model performance.
\end{enumerate}

The number of latent features, i.e. the reservoir dimension, is kept fixed to $H=50$ across all layers.
The weight matrices in each reservoir layer are initialized as described in Section~\ref{sec.RCgraphs}.
We set the convergence threshold $\epsilon=1e-5$ and we set the maximum number of iterations to $max_{iter} = 50$, meaning that \eqref{eq.iterated} is iterated until the norm of the difference between two consecutive updates is less than $1e-5$, or $50$ iterations are done.
We notice that, in practice, the convergence is almost always reached before surpassing the maximum number of allowed iterations.

The pooling methods do not depend on hyper-parameters, with two exceptions:
\begin{itemize}
    \item In \textit{NMF} pooling, it is possible to specify the number $K$ of super-nodes in the coarsened graph.
    For each graph
    $\mathcal{G}$, we let $K_\mathcal{G} = N_\mathcal{G}/2$, so that the number of vertices in a graph are halved each time.
    This is consistent with the two other pooling strategies, which also  reduce the number of nodes by a factor of 2.
    \item In Section~\ref{sec:ndp} we explained that a hyper-parameter $\delta$ controls the sparsification procedure in the \textit{NDP} pooling strategy.
    We set $\delta = 0.1$, meaning that all the edges with weight larger than $0.1$ are dropped.
    While this is a quite aggressive pruning, it allows to trade some accuracy in the graph representation with the computational efficiency of working with a graph that is sparse.
\end{itemize}

\bgroup
\def\arraystretch{1.2} 
\setlength\tabcolsep{.2em} 
\begin{table*}[!ht]
\footnotesize
\centering
\caption{
Results obtained for the Pyramidal architectures [\texttt{RGNN}, \texttt{Pool}]-[\texttt{RGNN}] and for the \textit{No-pool} architecture [\texttt{RGNN}]-[\texttt{RGNN}].
For each model, we report: i) the mean and the standard deviation of the classification accuracy obtained on the 5 external folds; ii) average training time (t\textsubscript{tr}) and test time (t\textsubscript{tr}), in seconds. 
We denote with ``Mem Err'' the experiments that failed due to memory errors.
}
\label{tab:results2}
\begin{tabular}{l|ccc|ccc|ccc|ccc}
\cmidrule[1.5pt]{1-13}
\multicolumn{1}{c}{\multirow{ 2}{*}{}} &  \multicolumn{3}{c}{\textbf{No-pool}} & \multicolumn{3}{c}{\textbf{Graclus}} & \multicolumn{3}{c}{\textbf{NMF}} & \multicolumn{3}{c}{\textbf{NDP}} \\
\multicolumn{1}{c}{} & 
\multicolumn{1}{c}{Acc.} & \multicolumn{1}{c}{t\textsubscript{tr}} & \multicolumn{1}{c}{t\textsubscript{ts}} & 
\multicolumn{1}{c}{Acc.} & \multicolumn{1}{c}{t\textsubscript{tr}} & \multicolumn{1}{c}{t\textsubscript{ts}} & 
\multicolumn{1}{c}{Acc.} & \multicolumn{1}{c}{t\textsubscript{tr}} & \multicolumn{1}{c}{t\textsubscript{ts}} & 
\multicolumn{1}{c}{Acc.} & \multicolumn{1}{c}{t\textsubscript{tr}} & \multicolumn{1}{c}{t\textsubscript{ts}} \\
\cmidrule[1.5pt]{1-13}
B-easy          & 96.9$\pm$0.6 & 8.9 & 2.3 & 93.4$\pm$1.4 & 6.5 & 1.6 & 87.6$\pm$1.2 & 6.1 & 1.5 & 91.5$\pm$0.5 & 5.3 & 1.3 \\
B-hard          & 75.3$\pm$2.0 & 8.5 & 2.1 & 68.7$\pm$2.9 & 5.6 & 1.4 & 62.1$\pm$1.9 & 4.7 & 0.6 & 67.9$\pm$2.7 & 4.4 & 0.3 \\
Mutagenicity    & 76.1$\pm$0.9 & 5.1 & 1.3 & 74.4$\pm$1.8 & 5.3 & 1.3 & 69.2$\pm$0.7 & 3.8 & 0.9 & 69.1$\pm$0.5 & 5.7 & 1.4 \\
Proteins        & 70.9$\pm$1.9 & 1.3 & 0.3 & 70.7$\pm$0.2 & 1.1 & 0.3 & 70.1$\pm$1.8 & 1.2 & 0.3 & 71.8$\pm$1.7 & 1.3 & 0.3 \\
Enzymes         & 45.5$\pm$2.9 & 0.6 & 0.1 & 37.0$\pm$0.3 & 0.5 & 0.1 & 33.5$\pm$2.4 & 0.4 & 0.1 & 36.7$\pm$4.5 & 0.4 & 0.1 \\
NCI1            & 70.4$\pm$0.8 & 5.0 & 1.3 & 68.1$\pm$0.8 & 5.3 & 1.3 & 66.9$\pm$2.1 & 4.2 & 1.0 & 67.4$\pm$1.2 & 4.5 & 1.1 \\
D\&D            & 76.4$\pm$1.7 & 8.4 & 2.2 & 72.5$\pm$0.9 & 8.2 & 2.0 & 71.4$\pm$1.7 & 9.4 & 2.3 & 71.2$\pm$3.0 & 8.2 & 2.0 \\
IMDB-BIN        & 71.3$\pm$1.9 & 0.7 & 0.2 & 70.7$\pm$5.1 & 0.9 & 0.2 & 63.2$\pm$2.3 & 0.8 & 0.1 & 69.7$\pm$1.9 & 0.9 & 0.2 \\
IMDB-MUL        & 49.5$\pm$2.3 & 0.9 & 0.2 & 48.5$\pm$1.9 & 1.3 & 0.3 & 46.1$\pm$1.7 & 0.9 & 0.2 & 47.3$\pm$2.7 & 1.1 & 0.3 \\
COLLAB          & 74.3$\pm$1.2 & 13.1 & 3.3 & 72.7$\pm$1.5 & 15.2 & 3.8 & 54.8$\pm$0.9 & 12.2 & 3.1 & 71.2$\pm$0.6 & 12.0 & 3.0 \\
Reddit-2K       & 84.7$\pm$0.9 & 31.5 & 8.0 & 81.6$\pm$0.9 & 41.7 & 10.6 & 52.2$\pm$4.1 & 37.5 & 9.5 & 80.5$\pm$2.0 & 13.0 & 3.2 \\
Reddit-5K       & 53.6$\pm$1.1 & 98.3 & 24.6 & \multicolumn{3}{c|}{Mem Err} & 35.3$\pm$1.4 & 98.5 & 24.7 & 49.3$\pm$1.3 & 42.5 & 10.6 \\
\cmidrule[1.5pt]{1-13}
\end{tabular}
\end{table*}
\egroup

In Table~\ref{tab:results2} we report the mean and the standard deviation of the classification accuracy obtained on the 5 external folds, obtained by the Pyramidal RGNN architecture with 2 blocks: [\texttt{RGNN}, \texttt{Pool}]-[\texttt{RGNN}], where \texttt{Pool} denotes one of the three pooling strategies: \textit{Graclus}, \textit{NMF}, or \textit{NDP}. 
For comparison, the same table also contains the results achieved by an architecture without pooling layers, \textit{No-pool}: [\texttt{RGNN}]-[\texttt{RGNN}].
For each model, we also report the mean training and test time (in seconds) obtained on the 5 folds. 
We indicate with ``Mem Err'' the experiments that failed due to a memory error.
Those errors occurred when the adjacency matrices of a certain dataset were very large or contained too many edges and it was not possible to store them in the 64GB of memory of the server used to perform the experiments.

Despite some variability, there are some patterns that emerge from the results in Table~\ref{tab:results2}.
As expected, \emph{No-pool}
generally
achieves the highest classification accuracy, with the only exception of Proteins on which the best accuracy is achieved in correspondence of \emph{NDP}. Among the Pyramidal RGNNN variants, the one using \emph{Graclus} generally leads to higher accuracy, and \emph{NDP} is runner up.
Regarding the computation times, on the other hand, Pyramidal RGNNs variants show a significant advantage, 
allowing calculation times to be reduced by up to almost 60\% compared to the No-pool model. The achievable speedup is particularly evident, e.g., in the case of the large Reddit datasets using \emph{NDP} pooling. Comparing the 3 Pyramidal RGNN variants, we notice that \emph{NDP} generally leads to the smallest times, and \emph{NMF} is runner up. \emph{Graclus}, on the other hand is almost always giving the worst results in terms of computational times,
in some cases being even slower than \emph{No-pool}.

\begin{figure*}
    \centering
    \includegraphics[width=\textwidth]{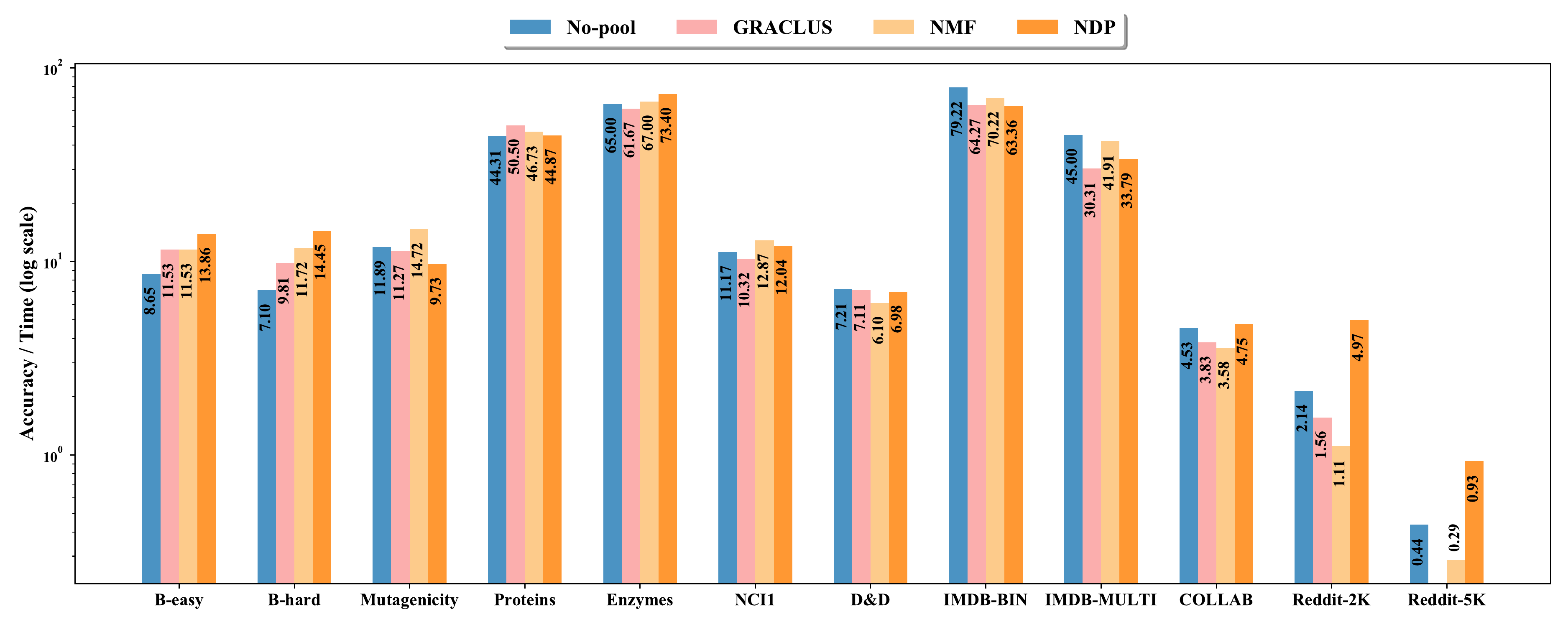}
    \caption{The performance measure $\frac{\text{classification accuracy}}{\text{train time + test time}}$ achieved by the Pyramidal architectures [\texttt{RGNN}, \texttt{Pool}]-[\texttt{RGNN}], compared to the results of the \emph{No-pool} architecture [\texttt{RGNN}]-[\texttt{RGNN}] on the 12 classification datasets.}
    \label{fig:acc_vs_time}
\end{figure*}

To summarize, \emph{No-pool} achieves a better classification accuracy, but the Pyramidal approach offers significant advantages in terms of computational complexity.
To evaluate with a single metric the trade-off between accuracy and complexity, we consider the ratio $\frac{\text{classification accuracy}}{\text{train time + test time}}$, which should be as high as possible.
Intuitively, this ratio tells us how many points of accuracy it is possible to achieve per second of computation.
In Figure~\ref{fig:acc_vs_time} we visualize in a bar plot the values of the ratio for the 4 architectures on the 12 classification datasets.
It is evident from the figure that the Pyramidal approach is preferable to \emph{No-pool} (with just a couple of exceptions for the IMDB datasets). In particular, Pyramidal RGNN with \emph{NDP} pooling shows the best overall accuracy/complexity trade-off, with outstanding results in both the Reddit datasets.

We continue the analysis of the classification performance by repeating the same classification experiment with an architecture composed by $L=3$ blocks, rather than $L=2$ blocks as in the previous case.
The results obtained by the Pyramidal architectures [\texttt{RGNN}, \texttt{Pool}]-[\texttt{RGNN}, \texttt{Pool}]-[\texttt{RGNN}] and by the \emph{No-pool} model [\texttt{RGNN}]-[\texttt{RGNN}]-[\texttt{RGNN}] are reported in Table~\ref{tab:results3}.

\bgroup
\def\arraystretch{1.2} 
\setlength\tabcolsep{.2em} 
\begin{table*}[!ht]
\footnotesize
\centering
\caption{
Results obtained for the Pyramidal architecture [\texttt{RGNN}, \texttt{Pool}]-[\texttt{RGNN}, \texttt{Pool}]-[\texttt{RGNN}] and for the \emph{No-pool} architecture [\texttt{RGNN}]-[\texttt{RGNN}]-[\texttt{RGNN}].
For each model, we report: i) the mean and the standard deviation of the classification accuracy obtained on the 5 external folds; ii) average training time (t\textsubscript{tr}) and test time (t\textsubscript{tr}), in seconds. 
We denote with ``Mem Err'' the experiments that failed due to memory errors.
}
\label{tab:results3}
\begin{tabular}{l|ccc|ccc|ccc|ccc}
\cmidrule[1.5pt]{1-13}
\multicolumn{1}{c}{\multirow{ 2}{*}{}} &  \multicolumn{3}{c}{\textbf{No-pool}} & \multicolumn{3}{c}{\textbf{Graclus}} & \multicolumn{3}{c}{\textbf{NMF}} & \multicolumn{3}{c}{\textbf{NDP}} \\
\multicolumn{1}{c}{} & 
\multicolumn{1}{c}{Acc.} & \multicolumn{1}{c}{t\textsubscript{tr}} & \multicolumn{1}{c}{t\textsubscript{ts}} & 
\multicolumn{1}{c}{Acc.} & \multicolumn{1}{c}{t\textsubscript{tr}} & \multicolumn{1}{c}{t\textsubscript{ts}} & 
\multicolumn{1}{c}{Acc.} & \multicolumn{1}{c}{t\textsubscript{tr}} & \multicolumn{1}{c}{t\textsubscript{ts}} & 
\multicolumn{1}{c}{Acc.} & \multicolumn{1}{c}{t\textsubscript{tr}} & \multicolumn{1}{c}{t\textsubscript{ts}} \\
\cmidrule[1.5pt]{1-13}
B-easy          & 97.6$\pm$0.8 & 10.1 & 2.7 & 85.0$\pm$1.8 & 7.6 & 1.9 & 76.0$\pm$1.2 & 6.3 & 1.6 & 83.6$\pm$0.8 & 5.5 & 1.4 \\
B-hard          & 76.4$\pm$1.5 & 9.8  & 2.6 & 54.4$\pm$1.6 & 5.9 & 1.5 & 49.6$\pm$1.5 & 6.3 & 1.6 & 57.4$\pm$1.8 & 4.8 & 1.2 \\
Mutagenicity    & 76.6$\pm$1.1 & 7.7  & 1.9 & 70.5$\pm$0.8 & 8.4 & 2.1 & 61.5$\pm$0.6 & 5.5 & 1.4 & 66.5$\pm$1.0 & 5.9 & 1.5 \\
Proteins        & 71.1$\pm$2.6 & 2.4  & 0.6 & 69.6$\pm$2.3 & 1.7 & 0.4 & 62.2$\pm$1.6 & 1.8  & 0.5 & 68.7$\pm$1.9 & 1.6 & 0.4 \\
Enzymes         & 47.2$\pm$4.5 & 0.8  & 0.2 & 30.3$\pm$2.7 & 0.7 & 0.2 & 27.8$\pm$2.7 & 0.6  & 0.2 & 29.6$\pm$2.5 & 0.6 & 0.1 \\
NCI1            & 70.5$\pm$1.9 & 6.4  & 1.6 & 65.0$\pm$1.3 & 6.9 & 1.7 & 61.2$\pm$1.2 & 4.1  & 1.0 & 63.4$\pm$1.4 & 4.3 & 1.1 \\
D\&D            & 73.7$\pm$1.2 & 13.5 & 3.4 & 71.2$\pm$3.8 & 9.8 & 2.4 & 63.0$\pm$3.3 & 11.1 & 2.8 & 69.1$\pm$2.7 & 9.6 & 2.3 \\
IMDB-BIN        & 72.3$\pm$1.4 & 1.3  & 0.3 & 70.8$\pm$2.3 & 1.3 & 0.3 & 57.2$\pm$2.0 & 0.8 & 0.2 & 68.3$\pm$1.4 & 1.2 & 0.3 \\
IMDB-MUL        & 49.7$\pm$0.8 & 1.4  & 0.4 & 47.5$\pm$2.9 & 1.2 & 0.3 & 42.0$\pm$1.6 & 1.1 & 0.3 & 47.0$\pm$2.5 & 1.2 & 0.3 \\
COLLAB          & 74.4$\pm$0.1 & 20.6 & 5.2 & 72.2$\pm$1.4 & 19.0 & 4.7 & \multicolumn{3}{c|}{Mem Err} & 69.2$\pm$0.5 & 13.9 & 3.4 \\
Reddit-2K       & 85.1$\pm$0.8 & 34.2 & 8.6  & 76.7$\pm$1.8 & 65.0 & 16.3 & \multicolumn{3}{c|}{Mem Err} & 74.7$\pm$1.0 & 10.7 & 2.7 \\
Reddit-5K       & 53.5$\pm$0.8 & 115.6 & 28.8 & \multicolumn{3}{c|}{Mem Err} & \multicolumn{3}{c|}{Mem Err} & 48.9$\pm$0.8 & 37.3 & 9.3 \\
\cmidrule[1.5pt]{1-13}
\end{tabular}
\end{table*}
\egroup

By comparing these new results with the previous ones for $L=2$ (in Table~\ref{tab:results2}), we notice that the \emph{No-pool} baseline
consistently achieves a slightly higher classification accuracy at the expense of a higher training and test time.
On the other hand, the classification accuracy of the Pyramidal architectures achieve a slightly worse classification accuracy when $L=3$, probably because too much information is discarded by two graph pooling operations.
In the architectures with \textit{Graclus} and \textit{NMF} pooling this drop in performance is not compensated by a speed-up in computing time.
For example, the Pyramidal architecture with $L=3$ and \textit{Graclus} achieves on Reddit-2K a worse classification accuracy and the train/test time is higher, compared to the architecture with $L=2$.
The situation is, however, different for the Pyramidal architecture with \textit{NDP} pooling, which achieves a significant improvement in computing time on Reddit-2K and Reddit-5K when $L=3$.
Once again, we highlight that on these two datasets the improvements in computing speed compared to the baseline are remarkable: \textit{NDP} is more than three time faster than \textit{No-pool}, as it can be trained in 37.3 seconds (on average) on Reddit-5K, rather than 115.6 seconds (on average).
Remarkably, those are the largest dataset and the improvement in the computing time here is crucial.

\begin{figure*}
    \centering
    \includegraphics[width=\textwidth]{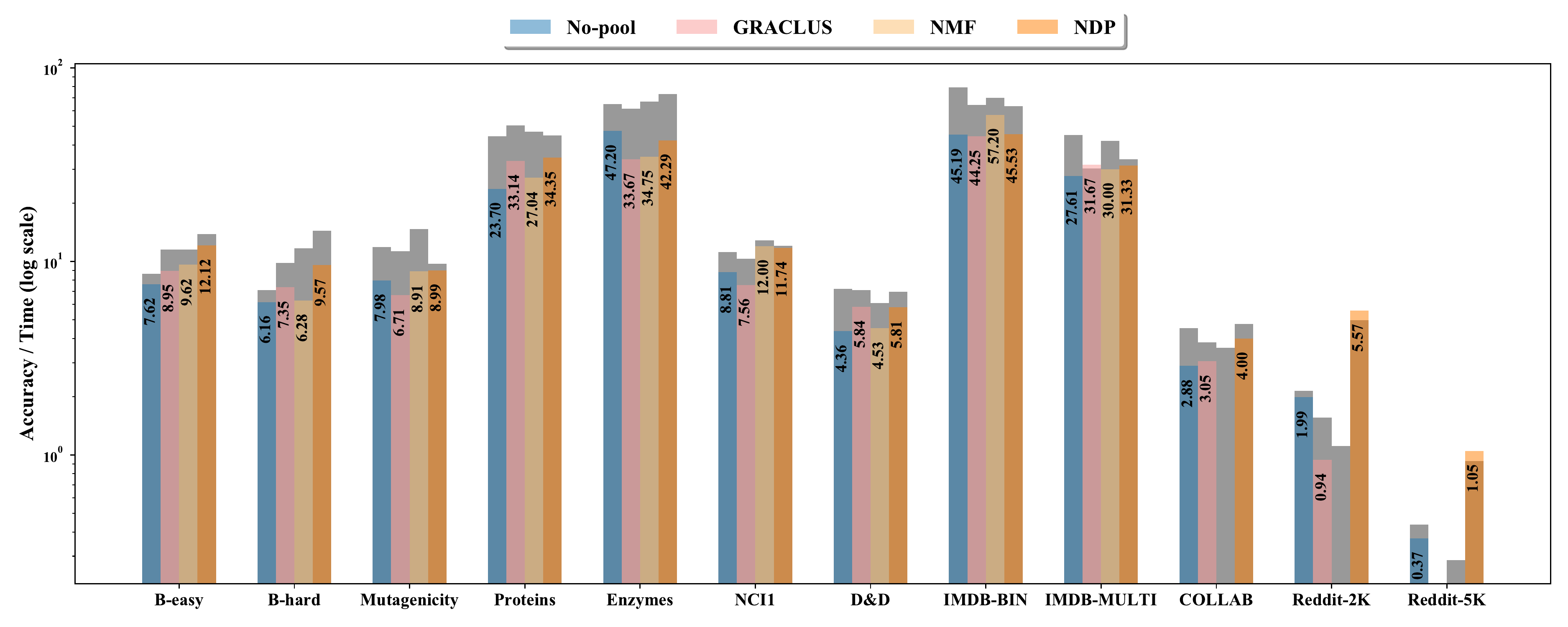}
    \caption{
    The performance measure $\frac{\text{classification accuracy}}{\text{train time + test time}}$ achieved by the Pyramidal and by the \textit{No-pool} architectures with $L=3$ blocks. For comparison, we depict in gray the results reported in Figure~\ref{fig:acc_vs_time}, i.e., those achieved by the architectures with $L=2$ blocks.}
    \label{fig:acc_vs_time_3L}
\end{figure*}

To effectively compare the differences in performance between the architectures with $L=2$ and $L=3$ blocks, we consider again the ratio between classification accuracy and the combination of train and test time.
In Figure~\ref{fig:acc_vs_time_3L} we report in colors the values of the ratio for $L=3$ and in gray the values of the ratio for $L=2$ (the same that were previously reported in Figure~\ref{fig:acc_vs_time_3L}).
Since the higher the ratio the better, it is immediately possible to notice that most architectures achieve better performance when $L=2$, confirming the observations made regarding Table~\ref{tab:results3}.
From Figure~\ref{fig:acc_vs_time_3L}
we can also notice that the Pyramidal architecture with \textit{NDP} and $L=3$ achieves a higher ratio on Reddit-2K and Reddit-5K.

To complete our analysis,
we link 
the achieved numerical results
to the outcomes of the theoretical investigation reported in Section~\ref{sec.analysis}. 
In particular, in  
Remark~\ref{remark.1} we have 
highlighted that the cost of graph embeddings computation in RGNN is an increasing function of the number of vertices ($N$), of the number of edges ($M$) and of the largest absolute eigenvalue of the normalized adjacency matrix ($\rho(\tilde \A)$).
In Table~\ref{tab:statistics_pooling}, we report these statistics for all the used datasets.

\begin{table*}[!ht]
\footnotesize
\centering
\caption{Statistics of the pooled graphs.
For each dataset, we report the average number of vertices ($N$), the average number of edges ($M$), and the average largest absolute eigenvalue of the normalized adjacency matrices ($\rho(\tilde \A)$). We detail the results corresponding to the original dataset, and after one (l1) ore two (l2) levels of the considered pooling algorithms. Missing values correspond to the memory errors in Table~\ref{tab:results2} and \ref{tab:results3}.}
\label{tab:statistics_pooling}
\begin{tabular}{l@{\hspace{1\tabcolsep}}r @{\hspace{1\tabcolsep}}r @{\hspace{1\tabcolsep}} r|@{\hspace{1\tabcolsep}}r @{\hspace{1\tabcolsep}}r @{\hspace{1\tabcolsep}} r|@{\hspace{1\tabcolsep}}r @{\hspace{1\tabcolsep}}r @{\hspace{1\tabcolsep}} r}
\cmidrule[1.5pt]{1-10} &
\multicolumn{3}{c}{\textbf{B-easy}}& 
\multicolumn{3}{c}{\textbf{B-hard}}&
\multicolumn{3}{c}{\textbf{Mutagenicity}}\\ 
     & \textbf{$N$} &  \textbf{$M$} &  \textbf{$\rho(\tilde \A)$}&
     \textbf{$N$} &  \textbf{$M$} &  \textbf{$\rho(\tilde \A)$}&
     \textbf{$N$} &  \textbf{$M$} &  \textbf{$\rho(\tilde \A)$} \\
\cmidrule[1.5pt]{1-10}
    \textbf{Original} & 
    147.8 & 922.6 & 1.00 & 148.3 & 572.3 & 1.00 & 30.3 & 91.8  & 1.00 \\
    \textbf{Graclus l1} &
    91.9 & 476.8 & 1.00  & 100.5 & 316.7 & 1.00 & 25.6 & 56.9 & 1.00\\
    \textbf{Graclus l2} &
    45.5& 209.3  & 1.00  & 50.1   & 154.3 & 1.00 & 12.8 & 33.7  & 1.00\\
    {\textbf{NMF l1}}& 
    73.0 & 3482.1 & 1.00 & 73.2 & 3110.7 & 0.99 & 14.9 & 223.4 & 1.00\\
    {\textbf{NMF l2}}&
    36.3 & 1388.9 & 1.00 & 37.0 & 1390.4 & 1.00 & 7.2 & 77.5 & 1.00\\
    \textbf{NDP l1}     & 73.0  & 7014.0 & 0.98 & 73.2 & 354.8  & 0.76     & 14.6  & 49.6 & 0.98 \\
    \textbf{NDP l2} & 36.1 & 359.0 & 0.97 & 37.5 & 170.1  & 0.66 & 6.7    & 26.7  & 0.92 \\ 
\cmidrule[1.5pt]{1-10} &
\multicolumn{3}{c}{\textbf{Proteins}}& 
\multicolumn{3}{c}{\textbf{Enzymes}}&
\multicolumn{3}{c}{\textbf{NCI1}}\\ 
     & \textbf{$N$} &  \textbf{$M$} &  \textbf{$\rho(\tilde \A)$}&
     \textbf{$N$} &  \textbf{$M$} &  \textbf{$\rho(\tilde \A)$}&
     \textbf{$N$} &  \textbf{$M$} &  \textbf{$\rho(\tilde \A)$} \\
\cmidrule[1.5pt]{1-10} 
\textbf{Original} 
& 39.1 & 72.8 & 1.00 & 32.6 & 62.1 & 1.00 & 29.9 & 62.1 & 1.00\\
{\textbf{Graclus l1}} & 25.9  & 95.0 & 1.00 & 22.4  & 79.2  & 1.00 & 21.0 & 56.1  & 1.00 \\
{\textbf{Graclus l2}} & 12.9 & 47.2 & 1.00 & 11.2  & 37.1  & 1.00 & 11.3  & 30.1 & 1.00 \\
{\textbf{NMF l1}}  & 19.3  & 580.5   & 1.00  & 16.1  & 238.4 & 0.99    & 15.1 & 193.2 & 1.00\\
{\textbf{NMF l2}}  & 9.5 & 220.1 & 1.00  & 7.8 & 75.8 & 1.00 & 7.1          & 62.9 & 1.00 \\
{\textbf{NDP l1}}& 19.2  & 106.0 & 0.98  & 16.2  & 83.3  & 0.98  & 15.2  & 44.8 & 0.98  \\
{\textbf{NDP l2}} & 9.5   & 56.4 & 0.89 & 8.1 & 38.9 & 0.97 & 7.4 & 23.1  & 0.96  \\ 
\cmidrule[1.5pt]{1-10} &
\multicolumn{3}{c}{\textbf{D\&D}}& 
\multicolumn{3}{c}{\textbf{IMDB-BINARY}}&
\multicolumn{3}{c}{\textbf{IMDB-MULTI}}\\ 
     & \textbf{$N$} &  \textbf{$M$} &  \textbf{$\rho(\tilde \A)$}&
     \textbf{$N$} &  \textbf{$M$} &  \textbf{$\rho(\tilde \A)$}&
     \textbf{$N$} &  \textbf{$M$} &  \textbf{$\rho(\tilde \A)$} \\
\cmidrule[1.5pt]{1-10} 
{\textbf{Original}}  & 284.3 & 715.6  & 1.00  & 19.7 & 96.5 & 1.00 & 13.0  & 65.9 & 1.00 \\
{\textbf{Graclus l1}} & 175.1    & 864.4  & 1.00    & 12.4 & 68.3  & 1.00 & 8.3 & 44.2 & 1.00\\
{\textbf{Graclus l2}} & 87.0 & 424.5 & 1.00 & 6.2 & 23.2 & 1.00 & 4.1      & 13.8  & 1.00 \\
{\textbf{NMF l1}}  & 141.3 & 24996.2  & 1.00 & 9.7  & 118.4 & 1.00         & 6.2 & 57.2  & 1.00 \\
{\textbf{NMF l2}}& 70.7  & 9625.0 & 1.00  & 4.6   & 27.3  & 1.00           & 3.0  & 13.4   & 1.00  \\
{\textbf{NDP l1}}  & 141.2 & 1364.7 & 0.98 & 9.7 & 86.7 & 0.99 & 6.3  & 46.8 & 0.96 \\
{\textbf{NDP l2}} & 70.2 & 902.2 & 0.99& 4.5  & 23.8 & 0.85 & 3.2 & 12.0  & 0.73 \\ 
\cmidrule[1.5pt]{1-10} &
\multicolumn{3}{c}{\textbf{COLLAB}}& 
\multicolumn{3}{c}{\textbf{Reddit-2K}}&
\multicolumn{3}{c}{\textbf{Reddit-5K}}\\ 
     & \textbf{$N$} &  \textbf{$M$} &  \textbf{$\rho(\tilde \A)$}&
     \textbf{$N$} &  \textbf{$M$} &  \textbf{$\rho(\tilde \A)$}&
     \textbf{$N$} &  \textbf{$M$} &  \textbf{$\rho(\tilde \A)$} \\
\cmidrule[1.5pt]{1-10} 
\textbf{Original}  & 74.5  & 2457.8   & 1.00   & 429.6  & 497.8   & 1.00     & 508.5  & 594.9& 1.00 \\
\textbf{Graclus l1} & 40.5  & 1589.3  & 1.00  & 1050.5 & 1157.6 & 0.99 & - & - & - \\
\textbf{Graclus l2} & 20.3  & 457.4 & 1.00 & 525.0 & 998.9 & 0.99 & - & - & - \\
\textbf{NMF l1} & 37.0 & 2338.2   & 1.00  & 214.5  & 90540.9  & 0.99       & 254.0  & 73675.5 & 1.00 \\
\textbf{NMF l2} & -   & -  & - & -  & -   & - & -  & - & -  \\
\textbf{NDP l1}   & 37.1 & 2284.1 & 0.89 & 142.1  & 1560.1   & 0.55 & 211.4  & 2268.4  & 0.28 \\
\textbf{NDP l2}   & 18.3  & 620.7 & 0.87 & 67.4 & 1091.4 & 0.69 & 103.2  & 1769.8 & 0.61 \\ 
\cmidrule[1.5pt]{1-10}
\end{tabular}
\end{table*}

We start by observing that all the pooling algorithms effectively reduce the number of vertices $N$ compared to the original graphs. Since the embedding cost in \eqref{eq.cost} depends on $N$, this is a first explanation of the consistent reduction in the required computation time observed for Pyramidal RGNN variants in comparison to \emph{No-pool}.
Among the pooling alternatives, we notice that \emph{NMF} and \emph{NDP} generally lead to a similar reduction in $N$, with the substantial difference that \emph{NDP} is easier to compute even on larger datasets (where memory errors are obtained in the other cases), and is then preferable.
Moreover, commenting the numerical results we noticed that in some cases \emph{Graclus} resulted to be slower than \emph{No-pool}. In Tables~\ref{tab:results2} and \ref{tab:results3} this is especially apparent for Reddit-2K.
Interestingly, we can see from Table~\ref{tab:statistics_pooling} 
that running \emph{Graclus} pooling on Reddit-2K tends to actually increase the 
number of vertices rather than decreasing it.
As discussed in Section~\ref{sec:graclus}, the reason is that  \textit{Graclus} needs to create a balanced binary tree by adding fake vertices before performing the pooling operation. 
In this way, the total number of vertices in the pooled graphs might increase in the intermediate layers and, in some cases, such an increment is significant, as observed in Table~\ref{tab:statistics_pooling}.
Even if the fake vertices are disconnected and do not affect the representations of the other vertices generated by the RGNN, their presence impact the computation time, as observed in Tables~\ref{tab:results2} and \ref{tab:results3}.

The models with \textit{Graclus} pooling generally achieve the highest classification accuracy among the Pyramidal RGNN models, but they are usually slower than the models equipped with \textit{NMF} and \textit{NDP} pooling, for the reason discussed above.
\textit{NDP} outperforms \textit{NMF} in terms of classification accuracy and usually achieves the lowest computational time among all methods.
This is particularly evident (Table~\ref{tab:results2}) in the largest datasets such as COLLAB, Reddit-2K and Reddit-5K, where \textit{NDP} is up to three times faster than the second fastest method.
By looking at Table \ref{tab:statistics_pooling}, we notice that the number of vertices ($N$) in the pooled graphs generated by \textit{NDP} and \textit{NMF} is similar but \textit{NDP} generates sparser graphs, i.e., with less edges and hence a smaller $M$. Since the embedding cost in \eqref{eq.cost} depends on $M$, this represents another computational advantage of Pyramidal RGNN with \emph{NDP} pooling.

By looking the computational times in Tables~\ref{tab:results2} and \ref{tab:results3}, it seems that the average number of vertices $N$ and the average number of edges $M$ do not completely explain the difference in the computational performance between the three pooling methods.
For example, on the COLLAB dataset, the \emph{Graclus} and \emph{NDP} methods result in a comparable reduction on N, while \emph{NDP} leads to larger values of M. Still, as shown in Table \ref{tab:results2} and especially Table \ref{tab:results3}, \emph{NDP} has lower computation times.
In Section~\ref{sec.analysis}, we discussed that the computational time of the 
embedding process
is heavily affected by the convergence speed of RGNN layers, which, in turn, depends on the largest absolute eigenvalue $\rho(\tilde{\mathbf{A}})$ (see Theorem~\ref{th.theorem} and Corollary~\ref{th.corollary}).

The columns of Table \ref{tab:statistics_pooling} with header $\rho(\tilde \A)$ report the average maximum eigenvalue 
(in absolute value) of the normalized adjacency matrices obtained with the different pooling methods.
We recall that the eigenvalues of a symmetrically normalized adjacency matrix $\tilde{\mathbf{A}} = \mathbf{D}^{-1/2}\mathbf{A}\mathbf{D}^{-1/2}$ are always contained in $[-1, 1]$.
We can see in Table~\ref{tab:statistics_pooling} that the coarsened graphs generated by \textit{NDP} have an average $\rho(\tilde \A)$
that is significantly smaller than  
what obtained in
the other cases, especially for the Reddit-2K and Reddit-5K datasets.
A smaller 
value of $\rho(\tilde \A)$
allows the iterative update in \eqref{eq.iterated} to converge faster, thus reducing the overall computational time.
The spectral properties of the coarsened graphs generated by \textit{NDP} derives from the Kron reduction procedure used to generate the new edges, which is described in detail in \cite{bianchi2019hierarchical}.

Overall, we see that \emph{NDP}  has intriguing features that generally allow us to impact all three factors that we identified as crucial to the computational cost of the graph embedding process in our theoretical analysis in Section \ref{sec.analysis}. Eventually, this leads to the advantageous performance trade-off highlighted in the experiments.

\subsection{Embedding visualization}
\label{sec.visualization}

In this section, we apply a dimensionality reduction technique for visualizing the structure of the graph embeddings computed by the architectures under consideration.
The main purpose of this analysis is to investigate if there are qualitative differences between the embeddings.
To perform dimensionality reduction, we adopt Linear Discriminant Analysis (LDA)~\cite{balakrishnama1998linear}, which computes low-dimensional projections by searching for linear combinations of the variables that best explain the data and the class labels.
LDA is a linear method and, therefore, the more the low-dimensional projections generated by LDA appear to be well separated, the better a linear classifier (like the readout used in the Pyramidal RGNN) can correctly predict the class of the graph embeddings.

The number of dimensions where LDA can project the data must be lower than the number of classes.
Therefore, a 2-dimensional visualization is possible only for datasets with 3 or more classes.
We selected B-Easy, B-Hard, Enzymes, and IMDB-MULTI datasets since are datasets with more than 2 classes.
Since we are performing a qualitative analysis, we do not perform model selection here, but we use the same hyper-parameters for every architecture ($\rho = 0.9$, $\omega_{in}=0.5$,  $\omega_{hid}=0.8$, $H=100$, $L=3$).

\begin{figure*}
    \centering
    \includegraphics[width=\textwidth]{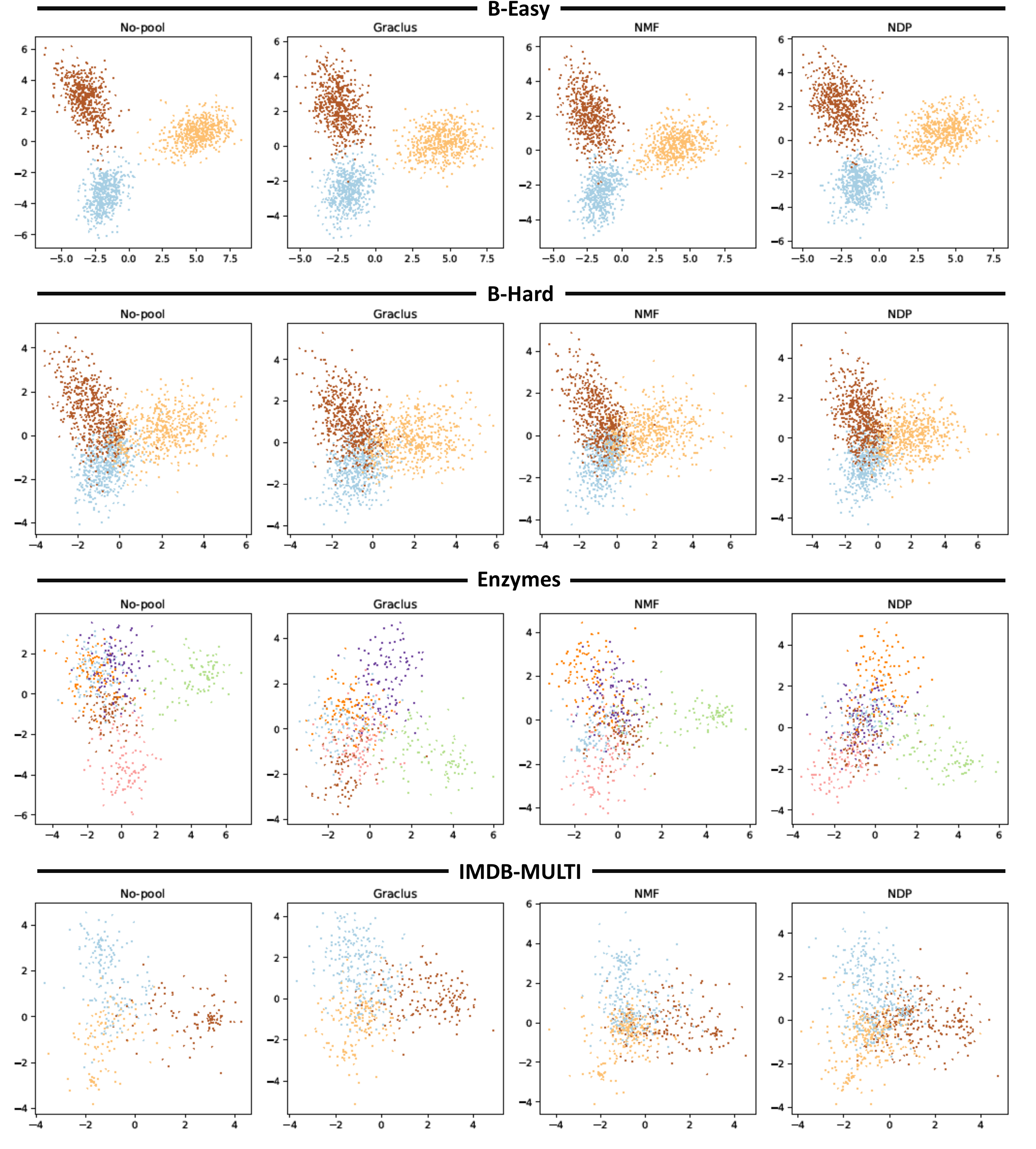}
    \caption{Visualization of the graph embeddings generated by the different architectures. The dimensionality reduction is performed with LDA.}
    \label{fig:visualization}
\end{figure*}

Figure~\ref{fig:visualization} shows the two-dimensional representations of the graph embeddings generated with the different architectures.
The first row shows the embeddings of the graphs in the B-Easy dataset.
As it can be seen in the figure, \textit{No-pool} manages to perfectly separate the three classes, while in the pyramidal architectures we observe some overlaps between the classes depicted in brown and in blue. 
The graph embeddings in the second rows, representing the graphs in the B-hard dataset, are more overlapping, suggesting that it is more difficult to classify correctly some of the graphs associated with the embeddings depicted in the middle.
The embeddings generated by \textit{No-pool} appear slightly more separated than those generated by the pyramidal architectures.
In the third row, we observe that some of the embeddings of the Enzymes dataset overlap considerably with each other.
The green class is the one that can be classified more accurately in every case.
Interestingly, some classes are separated better by different methods. 
For example, \textit{No-pool} separates well the pink class, \textit{Graclus} separates well the purple, and \textit{NDP} the orange class.
The last row shows the embeddings associated to the IMDB-MULTI dataset.
In this case, there is a larger difference between the methods under consideration.
The embeddings of \textit{No-pool} form some tight clusters, while the pyramidal methods form clusters that are less compact.
The embeddings generated by \textit{Graclus} appear more separated, while those generated by \textit{NMF} have a strong overlap.

The visualization results are well aligned with the numerical results reported in Table~\ref{tab:results3}: \textit{No-pool} yields graph embedding that can be separated slightly better by a linear classifier than those generated by the pyramidal variants.
Remarkably, apart from the occasional distinction mentioned above, results in Figure~\ref{fig:visualization} indicate that there are no strong qualitative differences between the graph embeddings generated by the different pyramidal architectures.
Hence, among the graph pooling methods, \textit{NDP} is preferable due to its lowest computational times, as analyzed in Section~\ref{sec.classification}.

\section{Conclusions}
\label{sec.conclusions}

In this paper, we introduced
a novel
Reservoir Computing model for graphs endowed with graph pooling operators.
In particular, we built a deep architecture with a pyramidal structure, where Reservoir 
GNN
layers that compute embeddings for the graph vertices are interleaved by graph pooling layers that generate coarsened representations of the underlying graph.
The vertex features in the last layer of the architecture are combined to generate a graph embedding vector, which can be used to perform tasks such as classification or dimensionality reduction.

The main advantage of the proposed pyramidal architecture is the reduction of the computing time required to generate the graph embeddings.
Therefore, our analysis focused on investigating the trade-off between performance in classification accuracy and computational complexity.
To this end, we introduced a formal mathematical analysis of the convergence of stable recursive graph embeddings and derived an upper-bound to its computational cost. Remarkably, we found that this bound depends on graph-theoretical properties of the input data, and in particular on the number of vertices, on the number of edges (hence on the density), and on the spectrum of the graph. By acting on these properties, the graph pooling algorithms can have a decisive impact on the computation times. Moreover, the derived bound allowed us to understand the differences in the computing time required by the different architectures analyzed.  The theoretical analysis presented in this paper is general and can be extended to other Reservoir Computing and recursive architectures for graphs.

Our experimental evaluation focused on two different types of tasks.
First, we performed graph classification and compared the classification accuracy and computing time of the proposed pyramidal architecture, equipped with different types of graph pooling operators, and of the same architecture without pooling layers.
Then, we considered a dimensionality reduction task, which allowed us to perform a qualitative analysis by visualizing the embeddings generated by the different models.

The experiments were performed on a large collection of datasets for graph classification.
Two of the datasets used in the experiments are introduced in this paper, to provide a solid benchmark to test the robustness of generic graph kernels or graph neural networks for classification.

Our results show that graph pooling allows to significantly accelerate the computing time at the price of partially reducing the quality of the graph embeddings.
We showed such a trade-off both from a quantitative and a qualitative perspective, i.e., in terms of classification accuracy and visualization of the graph embeddings.
Among the pooling methods, we found that Node Decimation Pooling offers the best trade-off between computational complexity and quality of the graph embeddings.

\bibliography{references}

\end{document}